\documentclass[lettersize,journal]{IEEEtran}
\usepackage{amsmath,amsfonts}
\usepackage{algorithmic}
\usepackage{algorithm}
\usepackage{array}
\usepackage[caption=false,font=normalsize,labelfont=sf,textfont=sf]{subfig}
\usepackage{textcomp}
\usepackage{stfloats}
\usepackage{url}
\usepackage{verbatim}
\usepackage{graphicx}
\usepackage{cite}
\usepackage{color}
\usepackage{multirow}
\usepackage{caption}
\usepackage{capt-of,etoolbox}
\usepackage{float}
\usepackage[table]{xcolor}

\definecolor{ccr}{RGB}{10,110,150}

\usepackage{hyperref}
\hypersetup{hypertex=true,
colorlinks=true,
linkcolor=ccr,
anchorcolor=ccr,
citecolor=ccr}

\begin{document}
\title{Propagating Sparse Depth via Depth Foundation Model for Out-of-Distribution Depth Completion}

\author{Shenglun Chen$^1$, Xinzhu Ma$^{2}$, Hong Zhang$^3$, Haojie Li$^3$, Zhihui Wang$^{1 \dagger}$, ~\IEEEmembership{Member,~IEEE}

\thanks{This work is supported in part by the National Natural Science Foundation of China (NSFC) under Grant (No.61932020), and the Taishan Scholar Program of Shandong Province (tstp20221128). $^{\dagger}$: corresponding author.}

\thanks{$^{1}$S. Chen and Z. Wang are with Dalian University of Technology, Dalian, China. {\tt\small shenglunch@gmail.com}, {\tt\small zhwang@dlut.edu.cn}.}

\thanks{$^{2}$X. Ma is with Beihang University, Beijing, China, {\tt\small xinzhuma@buaa.edu.cn}.}

\thanks{$^{3}$H. Zhang and H. Li are with the College of Computer Science and Engineering, Shandong University of Science and Technology, Qingdao, China. {\tt\small hongzh726@gmail.com, hjli@sdust.edu.cn}.}
}

\markboth{Journal of \LaTeX\ Class Files,~Vol.~14, No.~8, August~2021}%
{Shell \MakeLowercase{\textit{et al.}}: A Sample Article Using IEEEtran.cls for IEEE Journals}


\maketitle

\begin{abstract}
Depth completion is a pivotal challenge in computer vision, aiming at reconstructing the dense depth map from a sparse one, typically with a paired RGB image. Existing learning-based models rely on carefully prepared but limited data, leading to significant performance degradation in out-of-distribution (OOD) scenarios. 
Recent foundation models have demonstrated exceptional robustness in monocular depth estimation through large-scale training, and using such models to enhance the robustness of depth completion models is a promising solution.
In this work, we propose a novel depth completion framework that leverages depth foundation models to attain remarkable robustness without large-scale training.
Specifically, we leverage a depth foundation model to extract environmental cues, including structural and semantic context, from RGB images to guide the propagation of sparse depth information into missing regions. We further design a dual-space propagation approach, without any learnable parameters, to effectively propagates sparse depth in both 3D and 2D spaces to maintain geometric structure and local consistency. To refine the intricate structure, we introduce a learnable correction module to progressively adjust the depth prediction towards the real depth. 
We train our model on the NYUv2 and KITTI datasets as in-distribution datasets and extensively evaluate the framework on 16 other datasets. Our framework performs remarkably well in the OOD scenarios and outperforms existing state-of-the-art depth completion methods.
Our models are released in \href{https://github.com/shenglunch/PSD}{this link}.
\end{abstract}


\section{Introduction}
\IEEEPARstart{D}EPTH, a fundamental representation for perceiving the physical environment, is crucial in various computer vision tasks and applications, including object detection, 3D reconstruction, unmanned aerial vehicles, and robotic manipulation. While active depth sensors using technologies like structured light, time-of-flight, LiDAR, and radar provide reliable depth information, their inherent limitations give rise to sparse depth measurements, such as low spatial resolution and holes. In practical applications, a technique known as depth completion \cite{9286883} is commonly employed to enhance the density of measurements by using synchronized RGB images.

\begin{figure*}[t]
\centering
\includegraphics[width=1\linewidth]{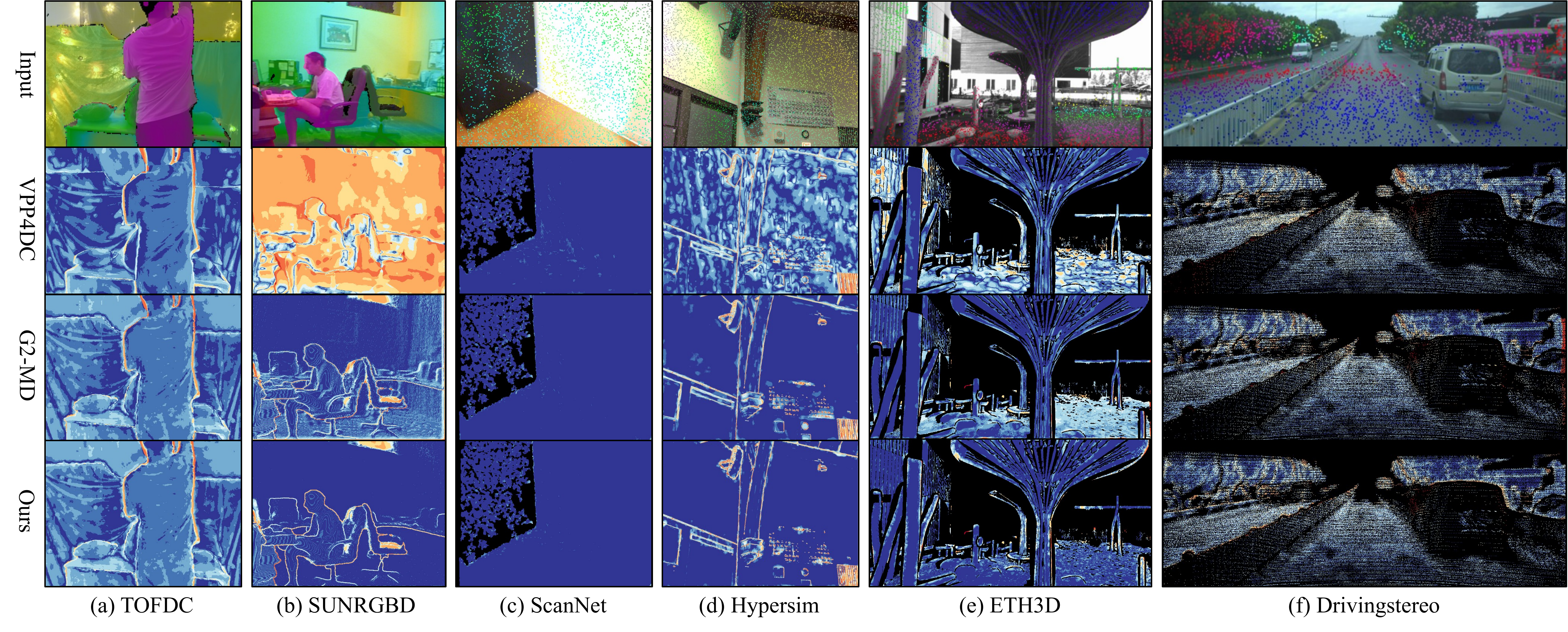}
\caption{{\bf Generalization of different methods.} We visualize the inputs (RGB images and sparse depth maps) and error maps (\textit{dark blue is better}) for VPP4DC \cite{bartolomei2023revisiting}, G2-MD \cite{10373158} and our model across six datasets: TOFDC \cite{yan2024tri}, SUNRGBD \cite{7298655}, ScanNet \cite{8099744}, Hypersim \cite{roberts2021hypersim}, ETH3D \cite{8099755}, and Drivingstereo \cite{yang2019drivingstereo}. The results highlight our model's superior generalization across diverse OOD scenarios.}
\label{fig_show}
\end{figure*}

Recent depth completion works \cite{zhao2021adaptive,zhang2023completionformer,Wang_2023_ICCV,kam2022costdcnet} have made significant progress, achieving impressive performance and fast inference speeds across specific datasets like NYUv2 \cite{SilbermanECCV12}, KITTI \cite{geiger2012we}, and VOID \cite{wong2020unsupervised}. However, most methods depend on carefully prepared in-distribution data during training. When applied to unseen scenarios, these methods suffer significant performance drops because the distribution of out-of-distribution (OOD) scenarios in scene appearance or depth characteristics differs from the training distribution. In monocular depth estimation that uses only RGB images to produce dense depth maps, several foundation models \cite{9178977,depthanything,depth_anything_v2} have shown exceptional robustness through large-scale dataset training. However, this solution is difficult to apply to depth completion, because the existing training samples of depth completion are limited and building new large-scale datasets is extremely challenging (due to the complicated data collection and synchronization process). In this case, leveraging strong pretrained depth foundation models to achieve generalized depth completion is a promising and valuable research.

Although monocular depth estimation also produce dense depth maps, the depth foundation models are challenging to be applied in depth completion, as they cannot recover the metric scale which is necessary for depth completion. Consequently, we do not directly use the depth maps generated by these foundation models in this work. In contrast, we calibrate the depth map as a spatial structural cue and combine it with semantic contextual cues from the model to propagate sparse depth to missing regions. As shown in Fig. \ref{fig_show}, our method predicts robust depth results in various OOD scenarios, which demonstrates the effectiveness of our designs.

Specifically, for depth completion models, we identify two main factors affecting the robustness: appearance characteristics from RGB images and depth characteristics from sparse depth measurements. Recent depth foundation models have effectively addressed various appearance characteristics. Thus, we mainly focus on extending these models to handle diverse depth characteristics without large-scale training. To this end, we propose PSD, a framework to \textbf{P}ropagate \textbf{S}parse \textbf{D}epth via strong depth foundation models for OOD depth completion.
In this framework, we leverage the depth foundation models to guide the propagation of sparse depth into missing regions to reconstruct a dense depth map.
To better propagate sparse depth cues, we also design a dual-space propagation approach that improves the 3D structure via propagating sparse depth in the 3D Euclidean space, and retains the local consistency in the 2D image space. 
Particularly, we convert the calibrated depth from the depth foundation model into a point cloud as the spatial structural cue to determine the neighbor regions using propagation. Additionally, we utilize the semantic contextual cue obtained from the model to propagate the sparse depth in the local region following \cite{cheng2018depth}. Without any extra learnable parameters, our dual-space propagation approach reconstructs an initial dense depth map.

Besides, although the initial depth exhibits promising performance, rivaling even the most advanced depth completion methods, unfortunately, distortions generated from the foundation model are also propagated, leading to a distorted scene structure. Therefore, we introduce a learnable correction module based on residual learning to improve the distorted scene structure. In this module, we estimate a possible residual range and refine this range for each individual pixel, using classification strategy to gradually adjust the final dense depth towards the real depth. 

Finally, we adopt NYUv2 \cite{SilbermanECCV12} and KITTI \cite{geiger2012we} datasets as the in-distribution data to train the proposed model and conduct extensive experiments. The results demonstrate that our framework exhibits impressive robustness and achieves state-of-the-art OOD performance across various scenes, including 16 real-world and synthesized datasets, such as VOID \cite{wong2020unsupervised}, Drivingstereo \cite{yang2019drivingstereo}, Hypersim \cite{roberts2021hypersim}, etc.

Overall, our contributions are summarized as follows:
\begin{itemize}

\item We investigate the effectiveness of leveraging pre-trained foundation models in depth estimation \cite{9178977,depthanything,depth_anything_v2,abs-2412-14015,abs-2410-02073,HuYZCLCWYSS24,abs-2502-20110} to enhance the capacities of depth completion, uncovering that the exceptional robustness of these models can be transferred into depth completion.

\item We propose a novel depth completion framework named PSD that effectively \textbf{P}ropagates \textbf{S}parse \textbf{D}epth into missing regions, guided by the spatial structure generated from the pre-trained depth foundation model.

\item We propose a dual-space propagation module that reconstruct the initial dense depth map in 3D and 2D spaces, and a correction module to refine the distorted structure. Both of them significantly improve the performance of our framework.

\item We conduct extensive experiments on 16 OOD datasets, which consistently demonstrate the effectiveness of the proposed framework.

\end{itemize}
\section{Related Work}
\subsection{Depth Completion}
Depth completion aims to reconstruct a dense depth map from the sparse one and the RGB image. Early works rely on interpolation \cite{camplani2012efficient}, stochastic models \cite{6619001}, morphological operators \cite{ku2018defense} or geometric models \cite{zhao2021surface}. 
Recently, significant advances have been made by leveraging convolutional neural networks to directly obtain the dense depth map. 
Ma \textit{et al}. \cite{ma2018sparse} first proposes a multi-modal fusion network to regress the dense depth map.
Following this, many works focus on improve the performance by deeply fusing multi-modal information \cite{liu2021fcfr,yan2021rignet,10413957,zhang2023completionformer}, embedding geometry structure \cite{Qiu_2019_CVPR,zhao2021adaptive,10377018}, introducing more priors \cite{9286883,9347719,8946876} and propagating sparse depth \cite{cheng2018depth,park2020non,10284921,Wang_2023_ICCV,yan2024tri,BP_Net}, \textit{etc}.

Among them, SCMT \cite{10413957} explicitly disentangles the hierarchical 3D scene-level features from the sparse depth and aggregates the sharp depth boundaries and object shape outlines into 2D features.
ACMNet \cite{zhao2021adaptive} converts the sparse depth to 3D point cloud as the input, and introduce the graph propagation to capture the
observed spatial geometry contexts.
GuideNet \cite{9286883} predicts content-dependent and spatially-variant kernels from the RGB image as a correlation prior to promote the multi-modal feature fusion.
Considering the sparse depth as a strong prior, CSPN \cite{cheng2018depth} creatively introduces the spatial propagation networks to propagate these sparse depth measurements to other regions, ultimately yielding outstanding performance.
Consequently, the propagation approach has been widely used in recent depth completion methods.
To address the limitations of static affinities and fixed neighborhood connections caused in conventional propagation approaches, DySPN \cite{10284921} refines the propagation process by dynamically adjusting affinity weights, propagation paths, and the number of neighbors.
To avoid directly convolving on sparse depth to generate the initial depth, BPNet \cite{BP_Net} proposes a multi-stage depth estimation block, which can propagate depth at the earliest stage and refine it in subsequent stages.

While these methods demonstrate promising results, they are limited to specific training datasets and struggle to generalize to OOD data. To address this challenge, we focus on robust depth completion, and propose a framework based on propagation that propagates the sparse depth in both 3D Euclidean and 2D image spaces, achieving astonishing generalization.

\subsection{OOD Depth Completion}
It is common to observe performance degradation when transferring a depth completion model trained on specific data to OOD testing data due to the domain gap.
VPP4DC \cite{bartolomei2023revisiting} is a pioneering work belongs to this research line. 
This method exploits the generalization capabilities of stereo matching networks to tackle depth completion, via the process of synthetic stereo pairs derived from a virtual pattern projection paradigm.
Leveraging the observation that the sparse depth modality exhibits a much smaller covariate shift than the RGB image, Park \textit{et al.} \cite{park2024testtime} propose an online test-time adaptation method that maintains a mapping from features encoding sparse depth solely to those encoding both RGB image and sparse depth, to align testing data to training data.
G2-MD \cite{10373158} tackles the challenge of OOD situations by collecting about 240K synthetic data, and represents it as `RGB+X' to accommodate RGB plus raw depth with diverse scene (scale/semantics), depth sparsity (0\%$\sim$100\%) and errors (holes/noises/blurs).
Park \textit{et al.} \cite{Park_2024_CVPR} propose a depth prompt module to allow the feature representation according to new depth distributions from either sensor types or scene configurations.

In this work, we adhere to the idea of propagation in the pursuit of better robustness and generalization, leading to superior performance across 14 different datasets.

\begin{figure*}[t]
\centering
\includegraphics[width=0.9\linewidth]{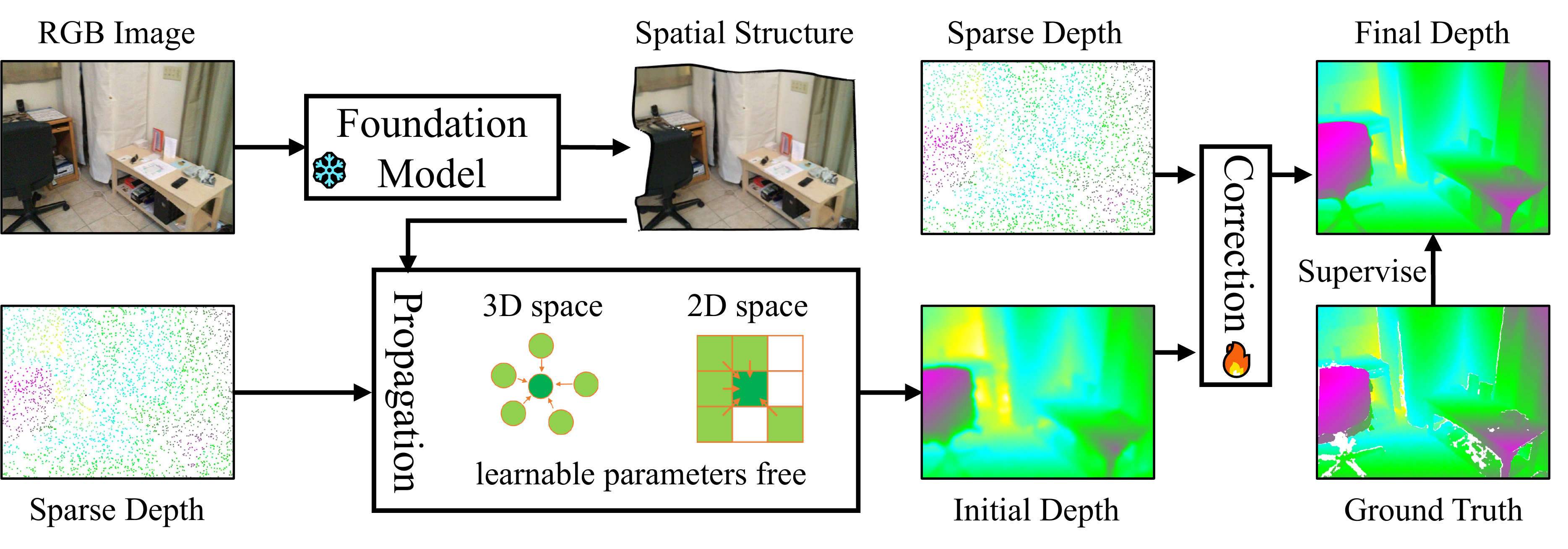}
\caption{{\bf The framework of PSD.} Given a pair of sparse depth map and RGB image, we first leverage a frozen foundation model to generate the spatial structure (Section \ref{sec_pc}). Then, with the guidance of the generated spatial structure, we propagate the sparse depth in both 3D and 2D spaces to obtain the initial depth (Section \ref{sec_diff}). Finally, we get the final depth by correcting the disrupted local structures (Section \ref{sec_corr}).}
\label{fig_frame}
\end{figure*}

\subsection{OOD Depth Estimation}
The purpose of depth estimation is to yield a depth map from the RGB image. ASN \cite{LongLL0TYW21, LongZZTLLZZW24} introduces a novel method that uses surface normal constraints and estimates similarity kernels to capture local geometry, achieving significant performance improvements. Recent efforts in OOD depth estimation focus on foundation models trained on large-scale data. 
MiDaS \cite{9178977} collects a mixed labeled datasets and proposes a robust training objective that is invariant to changes in depth range and scale to cover the data distribution.
Depth Anything \cite{depthanything} scales up the dataset by designing a data engine to collect and annotate large-scale unlabeled data, significantly enlarging the data coverage.
However, their depth predictions per pixel have no metric meaning and are only consistent relative to each other, limiting the applications, particularly in safety-critical domains such as robotics and autonomous driving.
ZoeDepth \cite{bhat2023zoedepth} incorporates distinct indoor and outdoor metric depth bins to fine-tune the MiDaS for metric depth estimation. 
Metric3D \cite{Yin_2023_ICCV} leverages the camera model to enhance metric depth estimation and introduces a canonical camera space transformation module to mitigate metric ambiguities stemming from diverse camera models in large-scale data. 
Depth Pro \cite{abs-2410-02073} aims to generate high-resolution, sharp depth maps with fine details for zero-shot metric monocular depth estimation. Depth Anywhere \cite{NEURIPS2024_e6c2e85d} and Depth Any Camera \cite{abs-2501-02464} both extend networks trained on perspective cameras to handle other cameras with different field-of-view angles. However, since RGB images lack real-world scale factors due to the perspective projection, metric depth estimation from a single RGB image is an ill-posed problem. This means these models cannot be directly used for depth completion.


We explore the effectiveness of foundational models for depth completion. Given that appearance and depth characteristics affect the robustness of depth completion, we focus on transferring the capability of them in representing appearance characteristics to address diverse depth characteristics.

\subsection{RGB-X Scene Understanding}
Depth completion, which generates dense depth maps from sparse depth maps and RGB images, is one of the critical tasks in RGB-X scene understanding. Recent works in other RGB-X tasks, such as 3D grounding \cite{MiyanishiAKK24}, RGB-D/Infrared object detection \cite{TIP.2025.3560488,YUAN2024102246}, RGB-D semantic segmentation \cite{10677963}, and RGB-Infrared depth estimation \cite{meng2025rgbthermalinfraredfusionrobust}, have demonstrated significant progress. Depth completion can further enhance the performance of other RGB-D tasks by providing more accurate depth information. For instance, in 3D object detection, both TED \cite{aaai.v37i3.25380} and SFD \cite{Wu_2022_CVPR} incorporate depth completion to improve detection accuracy. However, datasets from other tasks are typically not appropriate for training and fine-tuning depth completion methods due to the lack of corresponding sample pairs. Our framework exhibits superior generalization capability, potentially providing reliable depth information for these tasks without additional training or fine-tuning.

\section{Methodology}
\subsection{Problem Definition}
The goal of depth completion is to yield a dense depth map from a sparse one and synchronized RGB image. The conventional mathematical definition is as follows:
\begin{equation}
	\hat{D} = \Theta(S, I),
\end{equation}
where $\Theta$ denotes a depth completion network. $\hat{D} \in \mathbb{R}^{H\times W\times 1}$, $S \in \mathbb{R}^{H\times W\times 1}$, and $I \in \mathbb{R}^{H\times W\times 3}$ represent the dense depth, sparse depth, and RGB image, respectively. In this work, we propose a novel depth completion framework, trained on in-distribution data and demonstrating strong robustness when evaluated on OOD data.

\subsection{Overview}
Considering that sparse depth measurements are sufficiently reliable to serve as depth cues for neighboring regions lacking depth information \cite{cheng2018depth}, the proposed PSD framework propagates the sparse depth to nearby regions via the depth foundation model, as shown in Fig. \ref{fig_frame}.
Supported by depth foundation models \cite{9178977,depthanything,depth_anything_v2}, we propose a dual-space propagation approach that propagates sparse depth measurements in both 3D and 2D spaces. Specifically, we utilize the depth foundation model to generate spatial structural cues for propagation in 3D Euclidean space. Additionally, we use incidental semantic contextual cues to propagate sparse depth in 2D image space, following \cite{cheng2018depth}. Our proposed approach effectively maintains the 3D geometric structure and 2D local consistency of the scene. Finally, we introduce a learnable correction module to mitigate the disrupted structure caused by the propagation of distortions from depth foundation models.

\subsection{Spatial Structural Cue}
\label{sec_pc}
The conventional propagation approaches \cite{cheng2018depth,park2020non} restrict their neighbor search to the 2D space, owing to the absence of spatial structure. 
Recent depth foundational models \cite{9178977,depthanything} have shown significant robustness in monocular depth estimation, allowing us to promote propagation in 3D space by leveraging spatial structural cues from depth predictions.

\begin{figure}[t]
	\centering
	\includegraphics[width=0.9\linewidth]{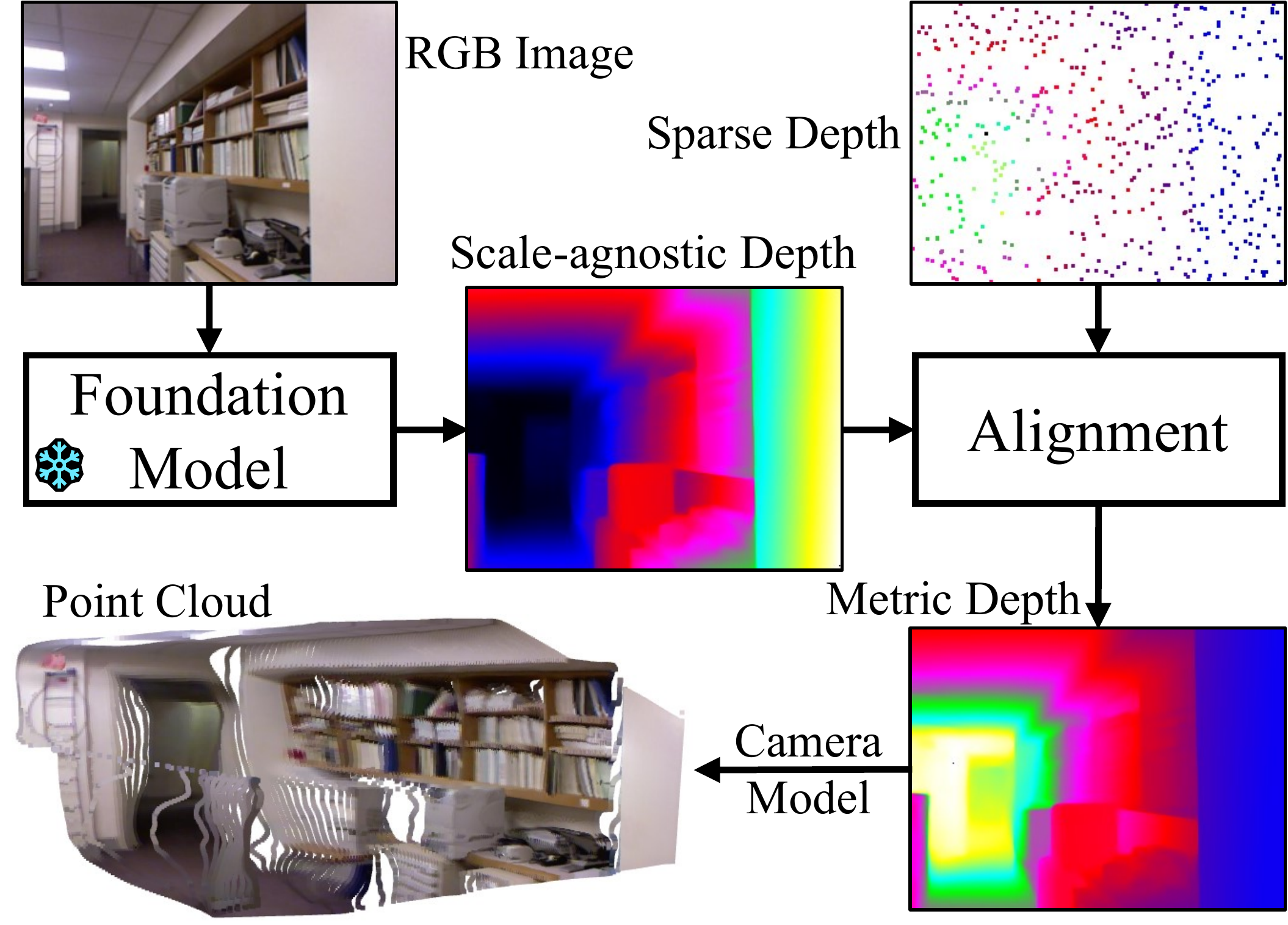}
	\caption{{\bf Generation of the spatial structure.} First, we use a frozen foundational model to derive the scale-agnostic depth. Then we align this depth with the sparse depth to obtain the metric depth. Finally, we back-project it into the 3D space to acquire the point cloud, which serves as the spatial structure.}
	\label{fig_align}
\end{figure}

Due to the perspective principle of imaging, depth foundational models generally face the challenge of intrinsic scale ambiguity. To address this issue, we introduce a scale alignment module to obtain a metric-scale depth map, which is then converted into a point cloud to represent the spatial structure.
As shown in Fig. \ref{fig_align}, we first leverage a frozen depth foundation model, like Depth Anything \cite{depthanything}, to obtain a scale-agnostic depth map from the RGB image. Next, we align this depth map and the sparse depth map using the least-squares method \cite{9178977}. The energy function associated with alignment is defined as follows:
\begin{equation}
E = \sum_{i\in \Omega} \left(\gamma D^{R}_i + \rho - S_i\right) ^ 2,
\label{eq_align}
\end{equation}
where $D^R$ denotes the relative depth with ambiguous scale, $\Omega$ denotes all valid measurements in the sparse depth, and $\gamma$ and $\rho$ are the scale and shift coefficients, respectively. 
It is important to note that the original scale-agnostic depth exists in the inverse depth space. To ensure numerical stability, we convert the sparse depth into the inverse depth space and then minimize the energy function. The resulting metric depth is then obtained as follows:
\begin{equation}
	D^M = \frac{1}{\gamma D^R + \rho}.
	\label{eq_dm}
\end{equation}
Finally, the metric depth is back-projected into 3D space using the camera's intrinsic parameters $K$ to obtain the point cloud $P$, as follows:
\begin{equation}
	P = D^MK^{-1}X,
\end{equation}
where $X$ denotes the homogeneous coordinates of 2D pixels. The point cloud is used as the spatial structure of the scene to facilitate propagation within the 3D space.

\subsection{Dual-Space Propagation}
\label{sec_diff}
To reconstruct a dense depth map, we design a dual-space propagation approach that propagates sparse depth measurements into missing regions in both 3D and 2D spaces.

\begin{figure}[t]
	\centering
	\includegraphics[width=1\linewidth]{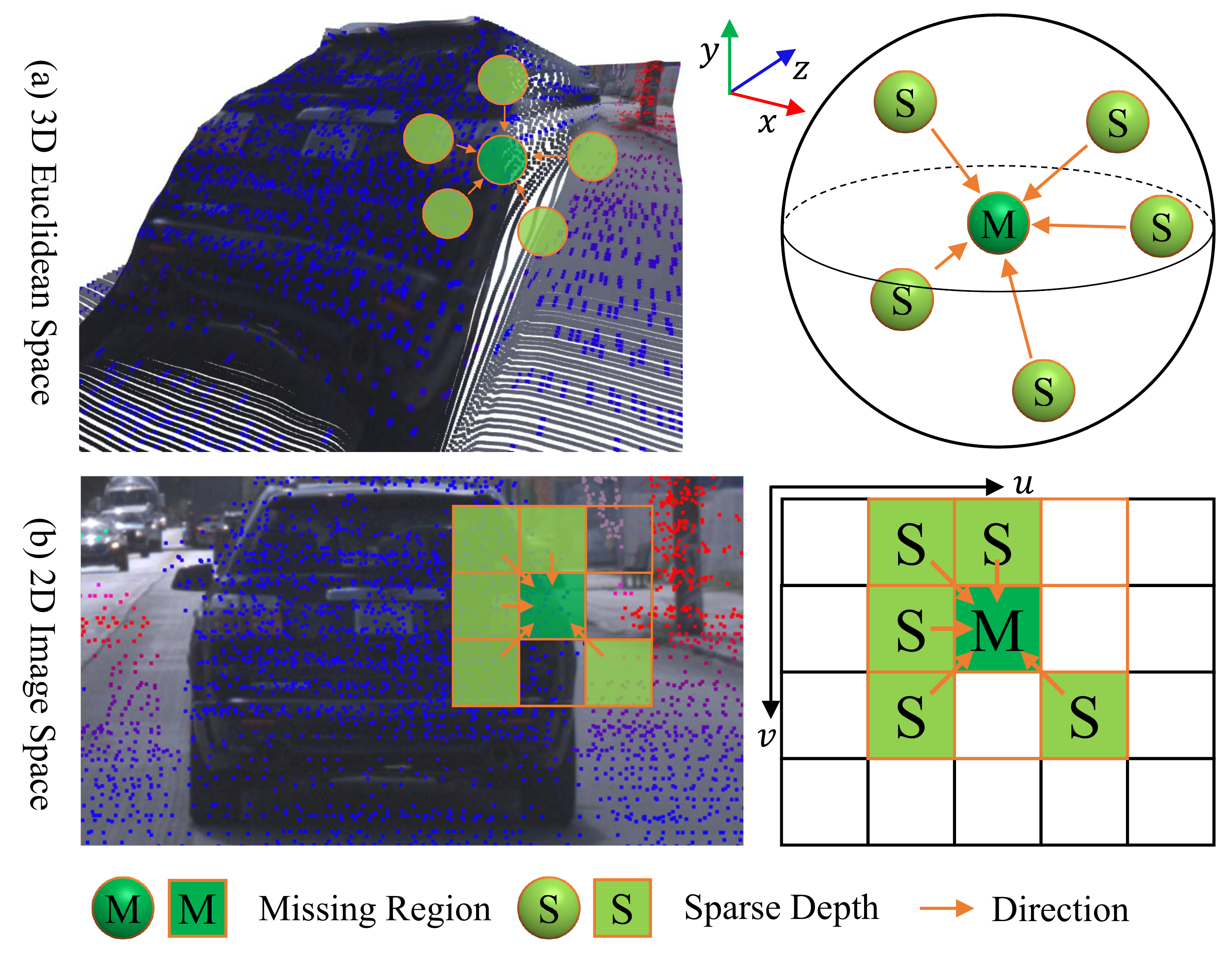}
	\caption{{\bf Dual-space propagation.} The proposed method propagates the sparse depth into missing regions based on both 3D and 2D geometrical constraints.}
	\label{fig_diff}
\end{figure}

For 3D space propagation, we use the spatial structure of the scene to determine neighboring regions and their respective affinities with the reference point. For a reference point lacking depth value, we select $k$ nearest neighboring points with sparse depth measurements. We use the Euclidean distance from a point to the reference point in the point cloud $P$ to determine whether it is a neighbor, and the far-away point may also be selected as a neighbor with a high sparsity.
For example, in Fig. \ref{fig_diff}, if measurements are missing near the taillight, the neighbors could be associated with the plate or even the road. To reduce the impact of irrelevant nearby neighbors, we utilize incidental semantic contextual cues to calculate their affinities to the reference point, as follows:
\begin{equation}
	A_i = \frac{F_i\times F_x^\top}{\sum_{j \in \mathcal{N}\left(x\right)} F_j \times F_x^\top},
\end{equation}
where $\mathcal{N}\left(x\right)$ denotes $k$ nearest neighbors of reference point $x$ in 3D space, $F \in \mathbb{R}^{1 \times C} $ denotes the feature of a point. All affinities are normalized in the neighboring region. Then, we aggregate the sparse depth $S$ of neighboring points to update the depth value of the reference point by:
\begin{equation}
	{D}^{3D}_x = \left(1-\eta\right)D^M_x + \eta \sum_{i \in \mathcal{N}\left(x\right)} A_i S_i,
\end{equation}
where $\eta \in \left[0, 1\right]$. Since $D^M$ is overly smooth, we replace it with a pre-filled depth using Gaussian filters \cite{ku2018defense}.

Previous propagation approaches generally employ a paradigm based on local consistency, where the depth values of two points on the same surface should be similar. They first propagate sparse depth measurements within local regions in the 2D space and then progressively increase the propagation range through iteration.
We adopt this paradigm to improve the local consistency of the dense depth map following \cite{cheng2018depth}:
\begin{equation}
	\begin{split}
		& A_i = \frac{F_i\times F_x^\top}{\sum_{j \in \mathcal{N}\left(x\right)} \left|F_j \times F_x^\top \right|} , \\
		& {D}^{2D}_x = A_x{D}^{3D}_x + \sum_{i \in \mathcal{N}\left(x\right)} A_i S_i,
	\end{split}
\end{equation}
where $\mathcal{N}\left(x\right)$ denotes neighbors of reference point $x$, which are the adjacent 8 pixels in 2D image space. $A_x$ denotes the affinity of the reference point, $A_x = (1-\sum_{i \in \mathcal{N}\left(x\right)} A_i)$. If a neighbor lacks a depth measurement, that is $S_i = 0$, we adopt the $D^{3D}_x$ for propagation.

Without introducing any additional learnable parameters, our dual-space propagation approach successfully propagates sparse depth measurements to reconstruct a dense depth map, referred to as the initial depth in Fig. \ref{fig_frame}. In Section \ref{sec_res}, we report on the zero-shot generalization capabilities of our dual-space propagation approach and the performance of different dual-space settings, including serial and parallel.

However, besides the sparse depth, the defects in depth foundation models are also propagated, such as the misinterpretation of complex structures, leading to distortion in the depth map. To address this problem, we introduce a correction module to refine the initial depth map.

\begin{figure*}[h]
	\centering
	\includegraphics[width=1\linewidth]{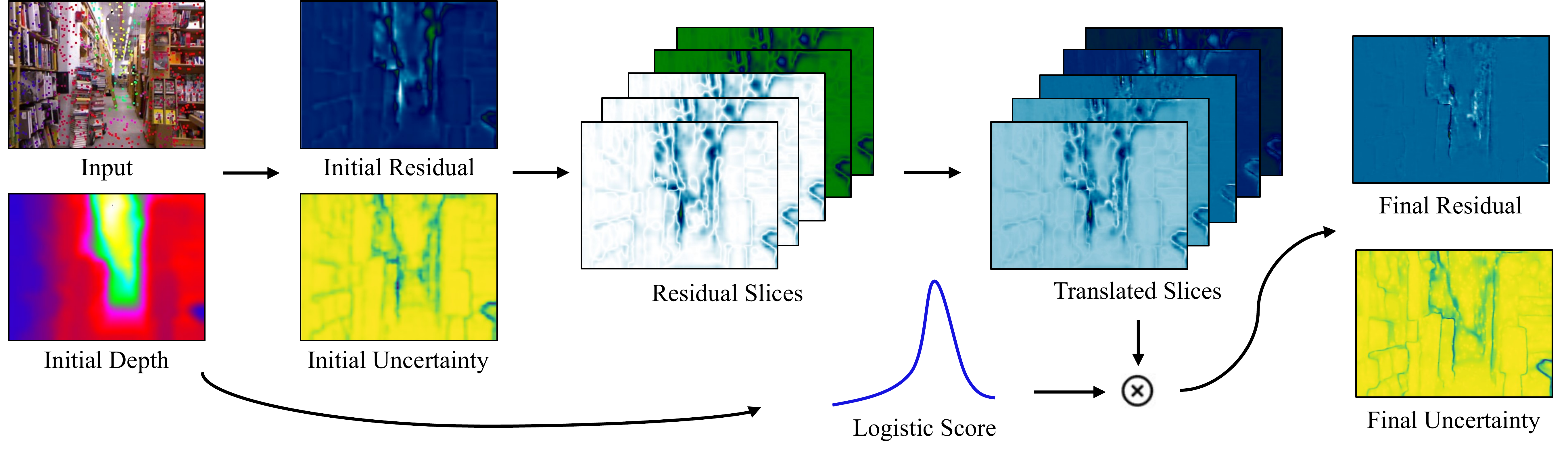}
	\caption{{\bf Correction module.} The module employs a classification strategy for progressive refinement. First, we estimate the initial residual and uncertainty. Next, we assign several possible residual slices around the initial result. Finally, we estimate the logistic score for each residual slice to calculate the final residual.}
	\label{fig_corr}
\end{figure*}

\subsection{Correction Module}
\label{sec_corr}
The proposed correction module is a learnable network module designed to progressively refine the initial depth prediction towards the real depth. As shown in Fig. \ref{fig_corr}, our correction module employs a classification strategy to estimate a residual map. Specifically, we first calculate the sparse residual between the sparse depth and the initial depth, which serves as prior information for residual estimation. We then feed this residual along with features extracted from the depth foundation model into a lightweight UNet to obtain an initial residual $R$ and an uncertainty $U$, which are further used to assign an individual residual range $[R_{min}, R_{max}]$ for each point, as follows:
\begin{equation}
	\begin{split}
		& R_{min} = \beta_{min} -(1 + \alpha_{min}) (1 + U) \left| R \right| , \\
		& R_{max} = \beta_{max} + (1 + \alpha_{max}) (1 + U) \left| R \right| , \\
	\end{split}
\end{equation}
where $\alpha$ and $\beta$ are learnable factors from zero. The uncertainty spans from 0 to 1. As the value approaches 1, it means that the residual is not sufficiently accurate, and we appropriately expand the range. We divide the range evenly into several discrete residual slices $\left\{R_i, 0 \le i \le n \right\}$, as follows:
\begin{equation}
	R_i = R_{min} + \frac{i\left(R_{max}-R_{min}\right)}{n},
\end{equation}
where $n$ is the number of residual slices. To enhance the continuity of the residual, we estimate a global offset to adaptively translate these residual slices:
\begin{equation}
	R_i^{'} = R_i + O, 
\end{equation}
where $O$ is the global offset. Then, we introduce a classification network based on UNet to estimate the logistic score for each residual slice and obtain the final residual by:
\begin{equation}
	\hat{R} = \sum_{i=0}^n L_i R_i^{'},
\end{equation}
where $L$ is the estimated logistic score normalized by $\rm{softmax}$ function. The final dense depth $\hat{D}$ is obtained by the adding final residual.

\subsection{Loss Function}
We train the proposed framework with a combination of L1 loss $L_1$ and uncertainty loss $L_{unc}$:
\begin{equation}
	Loss = L_1 + L_{unc},
\end{equation}
where $L_1$ is the L1 norm between the predicted depth and the ground truth, and $L_{unc}$ is the L1 norm related to uncertainty prediction, which encourages the prediction to approximate real uncertainty. However, since there is no ground truth for the uncertainty prediction, we model it with a function inspired by the probability density function of Laplace distribution \cite{10234651},
\begin{equation}
	U = 1 - \exp \left( - \frac{\left| \hat{D}-D \right|}{\tau \left( \hat{D}+D \right)} \right),
\end{equation}
where $\hat{D}$ denotes the predicted depth, $D$ denotes the ground-truth depth, $\tau$ is a coefficient that controls the tolerance for error, empirically set to 0.2.

\section{Experimental Setup}
\subsection{Datasets}
\noindent
\textbf{Training datasets.}
We use  NYUv2 \cite{SilbermanECCV12} and KITTI \cite{geiger2012we} as the in-distribution training data to train our PSD.
NYUv2 is acquired by the Microsoft Kinect sensor and comprises both depth and RGB sequences of indoor scenes.
The sparse depth is sampled from the depth map at a 1\% rate.
Following previous works \cite{yan2021rignet,10377018}, we adopt a subset of about 48K samples as the training set. 
KITTI is a large outdoor dataset captured by a driving vehicle. It provides about 86K semi-dense depth and RGB samples for training. 
The sparse depth is the raw output from the Velodyne LiDAR sensor, with a density of about 5\%.


\noindent
\textbf{Testing datasets.}
We evaluate the robustness of the proposed PSD framework across 16 OOD scenarios, including 12 real-world datasets: VOID \cite{wong2020unsupervised}, SUNRGBD \cite{7298655}, TOFDC \cite{yan2024tri}, DIML \cite{8359371}, Drivingstereo \cite{yang2019drivingstereo}, Argoverse \cite{8953693}, Cityscapes \cite{7780719}, DIODE outdoor \cite{vasiljevic2019diode}, DIODE indoor \cite{vasiljevic2019diode}, Middlebury \cite{ScharsteinS02}, ETH3D \cite{8099755}, ScanNet \cite{8099744}, 2 synthesized datasets: Hypersim \cite{roberts2021hypersim} and VKITTI2 \cite{cabon2020vkitti2}, a fisheye dataset: KITTI 360 \cite{LiaoXG23}, and a panoramic dataset: Stanford2D3D \cite{ArmeniSZS17}. The first 4 datasets contain the raw depth derived from keypoints obtained through odometry or captured by depth sensors such as structured light, and we directly input the raw depth into the network as the sparse depth. For the other datasets without raw depth, we use a fixed rate (1\%) to randomly sample the sparse depth from the ground truth.

\subsection{Evaluation Metrics}
TABLE \ref{tab_m} provides a summary of the evaluation metrics employed in our experiments, along with their respective units and definitions. These metrics include mean absolute error (MAE), root mean squared error (RMSE), mean absolute relative error (REL), and $\delta_\theta$ which represents the percentage of pixels that error is less than a threshold $\theta$.

\begin{table}[t]
\setlength\tabcolsep{12pt}
\centering
\caption{{\bf Summary of the metrics.}}
\begin{tabular}{l|ccc}
\hline
Metric     &  Unit & Definition & Better\\
\hline 
MAE  & mm  & $\frac{1}{\left| \Omega \right|} \sum_{p\in \Omega} \left| D_p - \hat{D}_{p} \right|$ & $\downarrow$\\
RMSE & mm  & $\sqrt{ \frac{1}{\left| \Omega \right|} \sum_{p\in \Omega} \left| D_p - \hat{D}_{p} \right|^2 }$ & $\downarrow$\\
REL  & \%  & $\frac{1}{\left| \Omega \right|} \sum_{p\in \Omega} \frac{\left| D_p - \hat{D}_{p} \right|}{D_{p}} $ & $\downarrow$\\
$\delta_\theta$ & \% & $\frac{1}{\left| \Omega \right|} \sum_{p\in \Omega} max \left(\frac{\hat{D}_{p}}{D_p}, \frac{{D}_{p}}{\hat{D}_p} \right)$ & $\uparrow$\\
\hline
\end{tabular}
\label{tab_m}
\end{table}

\begin{table*}[t]
\centering
\scriptsize
\caption{{\bf Quantitative results on 4 real-world datasets (with raw depth).} ALIGN/DS denotes the alignment and propagation approaches used in our PSD framework. N/K denotes the model trained on NYUv2/KITTI. Better: MAE $\downarrow$, RMSE $\downarrow$, REL $\downarrow$, $\delta_{1.25}$ $\uparrow$. \textbf{Best}, \underline{second best} results.}
\begin{tabular}{l|cccc|cccc|cccc|cccc}
\hline
\multirow{2}{*}{Method} & \multicolumn{4}{c|}{VOID \cite{wong2020unsupervised}} & \multicolumn{4}{c|}{SUNRGBD \cite{7298655}} & \multicolumn{4}{c|}{TOFDC \cite{yan2024tri}} & \multicolumn{4}{c}{DIML \cite{8359371}} \\ \cline{2-17} 
& MEA  & RMSE  & REL  & $\delta_{1.25}$ & MEA  & RMSE  & REL & $\delta_{1.25}$ & MEA  & RMSE  & REL & $\delta_{1.25}$ & MEA  & RMSE  & REL & $\delta_{1.25}$ \\ \hline
CFormer-N \cite{zhang2023completionformer} 
&119.1&217.5&8.77&85.51&7412&8621&52.3&12.67&845.1&864.4&46.0&2.444&951.2&1181&36.8&30.32 \\
CFormer-K \cite{zhang2023completionformer} 
&7527&8770&-&7.281&593.6&2068&10.5&96.77&48.70&126.5&2.92&98.62&858.0&1608&40.8&67.22 \\ 
\hline
IP\_Basic \cite{ku2018defense}
&186.9&408.5&12.8&86.73&702.2&2277&\underline{4.95}&96.43&150.2&298.0&7.86&92.57&694.0&1131&26.6&65.80 \\
VPP4DC-N	\cite{bartolomei2023revisiting}
&157.0&595.0&-&-&1772&2726&12.3&94.27&\textbf{34.78}&74.63&\textbf{2.10}&99.02&426.9&632.8&16.0&76.56 \\
VPP4DC-K	\cite{bartolomei2023revisiting}
&148.0&543.0&-&-&6104&31112&32.2&67.53&\underline{38.47}&89.23&\underline{2.27}&98.55&405.7&591.9&15.6&77.99 \\
G2-MD \cite{10373158}
&34.46&86.75&2.12&98.87&731.3&2423&9.72&96.56&40.48&77.32&2.42&99.09&431.6&660.6&17.2&76.44 \\
DP-N \cite{Park_2024_CVPR}	
&81.71&172.1&5.05&92.33&1096&3364&5.71&92.62&41.34&80.46&2.45&99.02&460.4&676.2&17.8&70.98 \\ 
DP-K \cite{Park_2024_CVPR}	
&987.9&1299&85.3&35.72&3058&3868&19.9&90.40&1085&1144&64.5&7.080&1174&1438&54.8&27.19 \\ 
PromptDA \cite{abs-2412-14015}	
&224.9&405.7&16.3&78.58&1149&2712&7.13&93.22&232.2&364.1&12.5&87.37&603.0&965.5&22.4&66.00 \\ 
\hline
ALIGN-D	
&88.04&153.6&5.25&97.30&1607&3018&15.4&93.75&64.46&98.25&3.85&98.94&\textbf{301.4}&\textbf{378.4}&\textbf{9.99}&\textbf{89.31} \\
ALIGN-Dv2
&87.61&157.6&5.15&97.54&1638&3146&15.7&93.16&74.11&109.6&4.47&98.32&\underline{336.8}&\underline{427.6}&\underline{11.2}&\underline{86.18} \\
ALIGN-M	
&117.4&187.0&7.27&95.16&1958&3390&17.6&90.17&83.71&118.2&4.95&98.24&341.1&433.5&11.4&85.35 \\
DS-D	
&47.33&107.3&3.05&98.10&471.3&1755&6.63&98.06&40.83&74.08&2.46&99.03&409.0&625.8&16.3&78.06 \\
DS-Dv2
&47.53&108.8&3.05&98.10&\underline{450.9}&{1690}&6.44&\textbf{98.15}&41.06&74.68&2.47&99.01&463.6&669.8&17.6&71.08 \\
DS-M	
&48.71&110.3&3.15&97.92&454.5&1701&6.49&\underline{98.14}&41.05&74.76&2.47&99.01&463.4&668.2&17.6&71.08 \\
\hline
PSD-N-D	
&\underline{28.39}&\textbf{75.04}&\underline{1.84}&\underline{99.05}&\textbf{380.0}&\textbf{1320}&\textbf{4.60}&97.72&39.92&\textbf{72.75}&2.40&\textbf{99.15}&392.9&585.2&15.3&78.99 \\ 
PSD-K-D	
&4891&5615&-&2.362&570.3&2030&7.95&97.59&45.98&107.9&2.78&98.77&829.2&1570&33.9&65.88 \\
PSD-NK-D	
&{29.71}&{76.20}&{1.98}&{99.02}&509.7&1767&7.17&97.56&40.11&76.78&2.41&\underline{99.14}&395.1&584.9&15.3&78.71 \\
PSD-NK-Dv2	
&36.27&87.26&2.79&98.58&523.1&1867&7.21&97.62&40.00&\underline{72.86}&2.41&\underline{99.14}&418.4&628.4&16.4&76.82 \\
PSD-NK-M	
&86.03&169.8&5.80&96.51&495.3&1728&6.47&97.52&40.98&85.12&2.45&99.13&407.4&597.9&15.8&77.21 \\
PSD-NK-PDA & 32.62 & 83.71&2.17& 98.70&525.8&1871&7.27&97.67&40.04&73.18&{2.41}&\underline{99.14}&436.6&696.8&17.2&{76.38} \\
PSD-NK-DPr & 34.67&91.45&2.57&98.38&519.1&1893&7.10&{97.73}&40.65&76.22&2.43&99.07&437.6&668.8&17.4&76.43  \\
PSD-NK-Mev2 &29.46&\underline{75.42}&1.89&99.01&472.8&1681&{6.12}&{97.74}&45.50&138.5&2.69&99.05&412.5&616.9&16.2&77.66 \\
PSD-NK-UDv2&\textbf{27.95}&75.51&\textbf{1.73}&\textbf{99.23}&{464.0}&\underline{1662}&6.25&97.67&40.85&86.76&2.45&99.12&404.6&603.3&{15.7}&78.04  \\
\hline
\end{tabular}
\label{tab_raw}
\end{table*}

\subsection{Implementation Details}
The proposed framework is trained for 45 epochs with 32 batch-size and AdamW optimizer. The learning rate is adjusted to \{1e-3, 7.5e-4, 5e-4, 2.5e-4, 1.25e-4, 5e-5, 1e-5\} at epoch \{0, 15, 20, 25, 30, 35, 40\}. We respectively integrate seven depth foundation models into our framework, including Depth Anything (D) \cite{depthanything}, Depth Anything v2 (Dv2) \cite{depth_anything_v2}, MiDaS (M) \cite{9178977}, PromptDA (PDA) \cite{abs-2412-14015}, DepthPro (DPr) \cite{abs-2410-02073}, Metricv2 (Mev2) \cite{HuYZCLCWYSS24}, and UniDepthv2 (UDv2) \cite{abs-2502-20110}.
During the training, we freeze the parameters of the embedded model for the entire duration. 

For a sample from NYUv2, we first downsample its size to 320 $\times$ 240, and then center-crop it to 304 $\times$ 228 to remove invalid regions. The corresponding sparse depth is randomly sampled from the ground truth using a fixed rate (1\%). Furthermore, we augment the training data through color jitter, horizontal flipping, random rotation, Gaussian noise \cite{10373158}, and pseudo-hole pattern \cite{9878677}.
For KITTI, since there are no valid depth measurements in the sky regions of the images, we crop the sample from the bottom center to a size of 1216 $\times$ 256.
The corresponding sparse depth is captured by LiDAR. The data augmentation includes color jitter, random cropping (832 $\times$ 256), and random mask for sparse depth. To integrate with NYUv2, we resize the KITTI samples to 304 $\times$ 228. The source code of this work will be publicly available for reproducibility.

Our dual-space propagation has no learnable parameters, enabling flexible reconfiguration during inference to enhance performance. For VKITTI2 \cite{cabon2020vkitti2}, we use parallel propagation in both 3D and 2D spaces, while other cases employ serial 3D-2D propagation. The number of $k$ nearest neighbors in 3D space is set as follows: 2 for Cityscapes \cite{7780719}, 3 for VKITTI2 \cite{cabon2020vkitti2}, 8 for TOFDC \cite{yan2024tri}, 12 for DIML \cite{8359371}, and 1 for others.

\section{Experimental Results}
\label{sec_res}
\subsection{Comparison with the State-of-the-Art Methods}
\label{sec_sota}
Our proposed framework exhibits remarkable performance in the OOD scenarios. 
To demonstrate this, we evaluate our model to the state-of-the-art methods on 16 OOD datasets.
For brevity, the models are named according to the following convention: PSD-\{DA\}-\{FM\}, where PSD is our framework, DA denotes the datasets used for training, and FM denotes the foundation model. We train and evaluate these models: PSD-N-D, PSD-K-D, PSD-NK-D, PSD-NK-Dv2, PSD-NK-M, PSD-NK-PDA, PSD-NK-DPr, PSD-NK-Mev2, and PSD-NK-UDv2, where N/K is NYUv2/KITTI training dataset. 

Furthermore, both our alignment and propagation components in the framework can be regarded as independent depth completion methods. In the experiment, we compare their robustness to others. The alignment approach (Section \ref{sec_pc}) is denoted as ALIGN. The dual-space propagation (Section \ref{sec_diff}) approach is denoted as DS.

\begin{table*}[t]
\centering
\scriptsize
\caption{{\bf Quantitative results on 4 real-world indoor datasets (without raw depth).} ALIGN/DS denotes the alignment and propagation approaches used in our PSD framework. N/K denotes the model trained on NYUv2/KITTI. Better: MAE $\downarrow$, RMSE $\downarrow$, REL $\downarrow$, $\delta_{1.25}$ $\uparrow$. \textbf{Best}, \underline{second best} results.}
\begin{tabular}{l|cccc|cccc|cccc|cccc}
\hline
\multirow{2}{*}{Method} & \multicolumn{4}{c|}{DIODE indoor \cite{vasiljevic2019diode} (Random 1\%)} & \multicolumn{4}{c|}{Middlebury \cite{ScharsteinS02} (Random 1\%)} & \multicolumn{4}{c|}{ETH3D \cite{8099755} (Random 1\%)} & \multicolumn{4}{c}{ScanNet \cite{8099744} (Random 1\%)} \\ \cline{2-17} 
& MEA  & RMSE  & REL  & $\delta_{1.25}$ & MEA  & RMSE  & REL & $\delta_{1.25}$ & MEA  & RMSE  & REL & $\delta_{1.25}$ & MEA  & RMSE  & REL & $\delta_{1.25}$ \\ \hline
CFormer-N \cite{zhang2023completionformer} 
&127.9&5299&3.89&\textbf{98.25}&381.0&1833&2.91&\textbf{97.30}&103.4&691.9&0.82&99.52&\textbf{12.20}&49.82&\textbf{0.63}&99.66 \\
CFormer-K \cite{zhang2023completionformer} 
&3468&4560&-&31.52&1016&2247&22.2&78.98&1049&1519&84.6&77.78&4109&4672&-&11.83 \\
\hline
IP\_Basic \cite{ku2018defense}
&208.4&1354&5.44&93.45&546.1&1578&5.83&91.58&135.2&610.3&2.69&97.69&49.82&160.3&2.49&97.80 \\
VPP4DC-N	\cite{bartolomei2023revisiting}
&96.61&1195&3.24&97.86&\underline{224.2}&\textbf{774.8}&2.89&97.02&59.91&231.3&0.74&\textbf{99.76}&14.05&53.35&0.74&99.59 \\
VPP4DC-K	\cite{bartolomei2023revisiting}
&128.7&1271&3.72&97.23&513.1&1497&5.34&92.78&193.2&746.2&1.99&98.65&21.78&75.05&1.12&99.17 \\
G2-MD \cite{10373158}
&115.9&1201&5.25&96.56&293.6&1001&3.56&96.04&115.7&376.8&1.64&98.50&13.42&50.78&0.69&\underline{99.67} \\ 
DP-N \cite{Park_2024_CVPR}	
&180.9&1620&5.85&97.86&321.6&884.7&3.33&96.54&115.8&418.7&1.03&99.29&15.21&52.45&0.79&99.64 \\ 
DP-K \cite{Park_2024_CVPR}	
&484.3&1461&22.0&77.64&835.8&1647&10.5&86.24&346.4&660.7&12.1&86.51&447.5&600.0&29.8&57.68 \\ 
PromptDA \cite{abs-2412-14015}	
&235.9&1466&6.67&93.88&832.5&1700&4.92&94.39&1217&2170&12.2&79.93&58.13&151.2&3.39&97.37 \\ 
\hline
ALIGN-D	
&327.5&1522&7.89&94.24&657.3&1199&7.81&94.93&551.5&1079&7.16&93.60&78.94&131.7&4.29&98.14 \\
ALIGN-Dv2
&324.7&1522&7.98&94.27&763.4&1271&7.80&94.76&472.7&920.9&6.39&95.01&77.31&135.8&4.13&98.14 \\
ALIGN-M	
&395.4&1641&10.1&91.60&1100&1776&11.6&87.53&758.5&1278&9.98&88.43&78.94&131.7&4.29&98.14 \\
DS-D	
&125.6&1173&4.10&97.65&387.1&930.3&5.16&94.52&88.63&307.6&2.14&98.17&23.41&63.80&1.28&99.39 \\
DS-Dv2
&125.6&1176&4.15&97.65&388.3&917.9&5.24&94.42&92.74&317.0&2.28&97.99&24.11&65.65&1.31&99.35 \\
DS-M	
&127.7&1182&4.18&97.60&395.4&937.4&5.28&94.29&80.35&287.5&1.84&98.56&23.94&65.20&1.30&99.38 \\
\hline
PSD-N-D	
&\textbf{94.49}&\textbf{1151}&\underline{2.86}&{98.00}&230.7&840.5&2.72&\underline{97.03}&32.26&{167.4}&{0.63}&99.61&{12.74}&\textbf{46.76}&{0.67}&\textbf{99.68} \\ 
PSD-K-D	
&679.0&1626&54.5&67.96&304.6&1113&3.29&95.84&676.1&899.3&58.8&84.57&1441&1649&-&30.05 \\
PSD-NK-D	
&97.30&1182&3.27&97.89&241.7&905.1&{2.66}&96.82&{29.27}&172.1&0.64&99.63&13.12&{49.45}&0.72&99.62 \\
PSD-NK-Dv2
&104.5&1203&3.54&97.81&258.4&999.6&\textbf{2.52}&96.87&\underline{26.03}&\textbf{136.7}&{0.55}&{99.69}&14.15&53.03&0.95&99.52 \\
PSD-NK-M	
&\underline{95.90}&{1165}&{3.19}&97.95&{225.8}&{825.0}&2.76&96.88&45.56&254.3&1.31&99.12&17.84&58.50&1.46&99.36 \\
PSD-NK-PDA&100.4&1181&3.29&97.94&237.1&906.8&\underline{2.59}&96.89&\textbf{25.40}&\underline{156.6}&\underline{0.54}&{99.71}&12.82&49.14&0.69&{99.65}  \\
PSD-NK-DPr&103.1&1193&3.28&97.70&303.5&1123&3.30&95.60&26.43&174.4&\textbf{0.51}&\underline{99.72}&13.75&54.94&0.82&99.48   \\
PSD-NK-Mev2&89.88&\underline{1162}&\textbf{2.81}&\underline{98.03}&235.9&901.6&3.13&96.55&155.2&885.7&6.56&98.34&\underline{12.23}&\underline{49.04}&\underline{0.65}&{99.65}  \\
PSD-NK-UDv2&205.7&1307&8.39&91.58&\textbf{219.8}&\underline{824.7}&2.76&{96.96}&35.20&205.2&0.91&99.57&{12.69}&50.45&{0.67}&99.64 \\
\hline 
\end{tabular}
\label{tab_rate2}
\end{table*}

\begin{table*}[t]
\centering
\scriptsize
\caption{{\bf Quantitative results on 4 real-world outdoor datasets (without raw depth).} ALIGN/DS denotes the alignment and propagation approaches used in our PSD framework. N/K denotes the model trained on NYUv2/KITTI. Better: MAE $\downarrow$, RMSE $\downarrow$, REL $\downarrow$, $\delta_{1.25}$ $\uparrow$. \textbf{Best}, \underline{second best} results.}
\begin{tabular}{l|cccc|cccc|cccc|cccc}
\hline
\multirow{2}{*}{Method} & \multicolumn{4}{c|}{Drivingstereo \cite{yang2019drivingstereo} (Random 1\%)} & \multicolumn{4}{c|}{Argoverse \cite{8953693} (Random 1\%)} & \multicolumn{4}{c|}{Cityscapes \cite{7780719} (Random 1\%)} & \multicolumn{4}{c}{DIODE outdoor \cite{vasiljevic2019diode} (Random 1\%)} \\ \cline{2-17} 
& MEA  & RMSE  & REL  & $\delta_{1.25}$ & MEA  & RMSE  & REL & $\delta_{1.25}$ & MEA  & RMSE  & REL & $\delta_{1.25}$ & MEA  & RMSE  & REL & $\delta_{1.25}$ \\ \hline
CFormer-N \cite{zhang2023completionformer} 
&5061&56523&9.36&94.77&27322&180269&38.5&79.66&6929&79467&15.6&92.89&2712&27427&18.9&85.02 \\
CFormer-K \cite{zhang2023completionformer} 
&756.4&2720&2.71&98.40&70.07&1127&0.16&99.90&3868&13644&17.8&92.09&2328&5919&30.5&78.14 \\
\hline
IP\_Basic \cite{ku2018defense}
&1136&3904&3.71&97.64&48.19&524.8&0.19&99.97&4745&17446&11.3&92.27&2249&6029&15.9&78.97 \\
VPP4DC-N	\cite{bartolomei2023revisiting}
&594.8&1989&2.37&98.94&901.8&3554&10.2&93.34&33545&46109&-&7.013&1754&5087&{14.7}&85.43 \\
VPP4DC-K	\cite{bartolomei2023revisiting}
&866.6&3153&3.11&98.06&217.1&1443&0.82&99.44&33050&43782&-&6.888&2386&6588&21.3&83.44 \\
G2-MD \cite{10373158}
&642.7&2274&2.12&98.98&121.6&341.5&0.53&\underline{99.98}&2731&11957&11.6&96.94&\underline{1579}&\underline{4749}&15.6&\underline{85.78} \\ 
DP-N \cite{Park_2024_CVPR}	
&1542&4597&4.55&95.57&75.86&751.0&0.28&99.84&4711&13763&15.7&89.37&2158&5586&17.9&82.56 \\ 
DP-K \cite{Park_2024_CVPR}	
&2142&4723&7.02&94.72&2356&4654&7.86&90.10&5531&14473&16.0&87.14&2634&5764&18.2&78.51 \\ 
PromptDA \cite{abs-2412-14015}	
&1835&5257&4.76&95.65&1497&3630&7.35&90.91&5501&17971&10.9&90.17&3185&8204&23.1&73.08 \\ 
\hline
ALIGN-D	
&2695&5649&8.25&92.81&3180&7340&9.85&90.32&30207&46298&90.5&0.298&5944&10556&32.8&55.14 \\
ALIGN-Dv2
&2494&5592&7.36&93.73&3394&7951&10.5&89.77&30150&46262&90.6&0.317&6194&11170&34.6&54.70 \\
ALIGN-M	
&3524&7307&10.3&87.69&3605&8036&11.3&87.05&30108&46087&90.2&0.308&5752&10303&32.0&54.96 \\
DS-D
&893.5&2502&2.77&98.56&174.0&720.7&0.59&99.95&3121&11083&12.2&93.62&1813&{4750}&17.5&84.64 \\
DS-Dv2
&886.0&2466&2.82&98.47&174.4&715.6&0.59&99.94&3628&11949&11.7&93.15&1835&4759&17.8&84.49 \\
DS-M	
&913.9&2579&2.79&98.49&173.1&721.8&0.58&99.95&3661&12030&11.8&93.00&1826&4778&17.7&84.56 \\
\hline
PSD-N-D	
&677.2&2458&2.01&99.21&17.97&239.4&0.07&\textbf{99.99}&2795&11686&\underline{9.64}&96.06&1661&4991&\textbf{14.0}&{85.49} \\ 
PSD-K-D	
&\textbf{521.3}&\textbf{1591}&\textbf{1.67}&\textbf{99.53}&\textbf{12.29}&{180.0}&\textbf{0.05}&\textbf{99.99}&\textbf{2422}&\textbf{10606}&{10.9}&\textbf{97.24}&1725&5180&17.9&84.56 \\
PSD-NK-D	
&556.3&{1655}&1.78&99.45&15.18&217.9&\underline{0.06}&\textbf{99.99}&2460&10778&11.6&97.15&1628&4996&16.9&{85.49} \\
PSD-NK-Dv2	
&553.1&1667&1.78&99.44&15.44&238.7&\underline{0.06}&\textbf{99.99}&2481&10828&12.0&97.09&1636&4983&15.6&84.86 \\
PSD-NK-M	
&{541.3}&1667&{1.74}&\underline{99.49}&{14.21}&\underline{176.5}&\underline{0.06}&\textbf{99.99}&\underline{2432}&{10740}&11.8&\underline{97.20}&{1591}&4875&15.8&85.15 \\
PSD-NK-PDA &549.3&1699&1.73&99.48&13.57&202.1&\textbf{0.05}&\textbf{99.99}&2637&10927&9.78&96.85&1702&5045&14.8&83.77 \\
PSD-NK-DPr &543.0&1828&\underline{1.72}&99.44&\underline{13.01}&\textbf{171.8}&\textbf{0.05}&\textbf{99.99}&2519&10893&10.0&96.96&1713&5136&\underline{14.4}&83.80  \\
PSD-NK-Mev2 &\underline{533.4}&\underline{1600}&1.73&\underline{99.49}&14.26&219.2&\underline{0.06}&\textbf{99.99}&2587&\underline{10722}&\textbf{9.40}&96.86&\textbf{1503}&\textbf{4729}&14.7&\textbf{86.30} \\
PSD-NK-UDv2 &647.9&2043&2.21&99.01&82.52&729.7&0.27&{99.66}&2478&10968&12.1&97.12&3430&8382&19.1&77.15  \\
\hline 
\end{tabular}
\label{tab_rate}
\end{table*}

\begin{table*}[t]
\centering
\scriptsize
\caption{{\bf Quantitative results on 4 challenging datasets (without raw depth).} ALIGN/DS denotes the alignment and propagation approaches used in our PSD framework. N/K denotes the model trained on NYUv2/KITTI. Better: MAE $\downarrow$, RMSE $\downarrow$, REL $\downarrow$, $\delta_{1.25}$ $\uparrow$. \textbf{Best}, \underline{second best} results.}
\begin{tabular}{l|cccc|cccc|cccc|cccc}
\hline
\multirow{2}{*}{Method} & \multicolumn{4}{c|}{Hypersim \cite{roberts2021hypersim} (Random 1\%)} & \multicolumn{4}{c|}{VKITTI2 \cite{cabon2020vkitti2} (Random 1\%)} & \multicolumn{4}{c|}{KITTI 360 \cite{LiaoXG23} (Random 1\%)} & \multicolumn{4}{c}{Stanford2D3D \cite{ArmeniSZS17} (Random 1\%)}\\ \cline{2-17} 
& MEA  & RMSE  & REL &$\delta_{1.25}$ & MEA  & RMSE  & REL & $\delta_{1.25}$ & MEA  & RMSE  & REL &$\delta_{1.25}$ & MEA  & RMSE  & REL &$\delta_{1.25}$\\ \hline
CFormer-N \cite{zhang2023completionformer} 
&245.8&1981&{1.29}&99.02&7729&93270&14.0&91.02&2870&16453&18.0&69.18&\textbf{18.14}&\textbf{147.1}&\textbf{1.66}&\textbf{99.35} \\
CFormer-K \cite{zhang2023completionformer} 
&1920&3243&-&61.97&3426&12455&15.0&90.70&7322&10265&-&40.77&5295&6562&-&7.288  \\
\hline
IP\_Basic \cite{ku2018defense}
&244.9&868.6&3.18&96.95&3382&12890&8.64&91.24&3341&7246&23.1&69.56&72.18&313.7&3.60&97.67 \\
VPP4DC-N	\cite{bartolomei2023revisiting}
&109.3&396.1&2.46&98.81&\textbf{1969}&9070&\textbf{5.34}&\underline{96.27}&1439&3988&13.5&87.77&24.95&183.5&2.20&98.90 \\
VPP4DC-K	\cite{bartolomei2023revisiting}
&275.0&4273&3.63&97.92&2509&9748&8.33&94.77&4213&115197&75.0&73.97&46.73&275.7&2.88&97.99 \\
G2-MD \cite{10373158}
&160.0&448.1&1.33&\textbf{99.28}&2124&9523&7.67&\textbf{96.33}&7206&11147&58.0&18.43&483.4&780.4&29.2&59.63  \\ 
DP-N \cite{Park_2024_CVPR}	
&256.5&658.8&1.70&98.65&4127&11935&11.6&88.37&8726&12802&-&1.923&36.68&164.8&2.81&99.06  \\ 
DP-K \cite{Park_2024_CVPR}	
&722.6&1293&22.1&83.54&4294&11673&11.9&87.61&9435&13785&67.4&8.452&1806&2110&96.2&1.019  \\
PromptDA \cite{abs-2412-14015}	
&1589&2670&5.94&92.66&4486&14680&7.69&90.44&3561&7386&23.4&61.01&100.8&320.2&5.69&96.07 \\  
\hline
ALIGN-D	
&572.8&973.3&8.21&92.51&8321&21070&18.3&79.59&2512&5393&17.0&78.45&440.4&792.3&23.3&58.83  \\
ALIGN-Dv2
&476.2&828.9&5.42&96.93&6880&18350&16.3&84.34&2577&5526&17.2&77.39&465.4&830.9&25.2&56.55  \\
ALIGN-M	
&797.5&1310&10.4&88.14&9546&23094&23.6&72.48&3092&6601&20.4&71.29&457.0&827.2&24.7&57.57   \\
DS-D	
&133.5&413.3&1.79&98.53&2845&9175&10.1&93.30&2048&4707&12.9&81.24&49.72&210.8&2.83&98.52  \\
DS-Dv2	
&131.4&406.8&1.77&98.57&2633&9049&7.96&94.33&1988&4629&12.6&81.92&49.49&209.5&2.84&98.52  \\
DS-M	
&140.5&433.7&1.84&98.47&2710&9632&8.10&93.95&2018&4662&12.8&81.61&50.52&214.5&2.86&98.46  \\
\hline
PSD-N-D	
&85.06&{377.6}&\underline{1.25}& {98.98} &2279&8843&7.04&94.89&1550&4896&8.42&90.40&\underline{20.04}&\underline{159.6}&\underline{1.71}&\underline{99.29} \\ 
PSD-K-D	
&316.5&765.3&27.8& 86.36  & \textbf{2059}&\underline{8564}&{6.88}&95.40&1883&4370&24.1&81.07 &1725&2414&-&19.61 \\
PSD-NK-D	
&{77.84}&389.8&2.47& 98.69  &{2068}&\textbf{8430}&6.91&95.30&2670&6049&25.4&78.08&24.32&170.8&2.60&98.93 \\
PSD-NK-Dv2	
&{75.66}&{377.0}&3.43& 98.71& 2135&8799&8.07&{95.41}&1165&\textbf{3502}&\textbf{7.66}&92.32&22.73&{169.6}&2.34&99.11  \\
PSD-NK-M	
&81.56&381.4&5.07& 98.66 & 2117&8775&7.60 &95.36  &2385&4670&32.2&74.01&29.57&190.4&3.33&98.75       \\
PSD-NK-PDA&\underline{66.06}&\underline{352.6}&2.25&98.89&2146&8731&{6.28}&95.32&\underline{1145}&\underline{3528}&\underline{7.73}&\textbf{92.60}&{22.18}&176.8&{2.04}&{99.21} \\
PSD-NK-DPr &79.38&406.4&4.69&98.42&2134&9007&6.72&95.19&1193&3665&8.01&91.85&24.90&190.5&2.19&98.87  \\
PSD-NK-Mev2 &\textbf{60.38}&\textbf{347.8}&\textbf{1.06}&\underline{99.07}&2174&9364&\underline{5.55}&{95.49}&23837&29488&-&22.19&244.1&618.4&23.9&81.78 \\
PSD-NK-UDv2 &853.7&1925&17.2&82.07&2136&8741&8.07&95.14&\textbf{1143}&3641&\textbf{7.66}&\underline{92.57}&25.60&175.4&{2.14}&99.18  \\ \hline
\end{tabular}
\label{tab_syn}
\end{table*}

\noindent
\textbf{Results on real-world datasets (with raw depth).}
TABLE \ref{tab_raw} reports the OOD performance on real-world datasets, where raw depth is used as sparse depth. Data augmentation struggles to simulate real sparse patterns, posing a challenge for robust depth completion in practical applications. Our proposed PSD exhibits substantially superior performance compared to previous methods, particularly VOID \cite{wong2020unsupervised} and SUNRGBD \cite{7298655} datasets. On TOFDC \cite{yan2024tri}, our PSD-N-D and PSD-NK-Dv2 achieve comparable performance to VPP4DC \cite{bartolomei2023revisiting}.
For DIML \cite{8359371}, which serves as a pre-training dataset for most depth foundation models, the alignment approach ALIGN-D achieves the best results. However, since the raw depth data contains noise, our DS propagation approach still propagates it, leading to a decline in performance.

\begin{figure*}[t]
\centering
\includegraphics[width=1\linewidth]{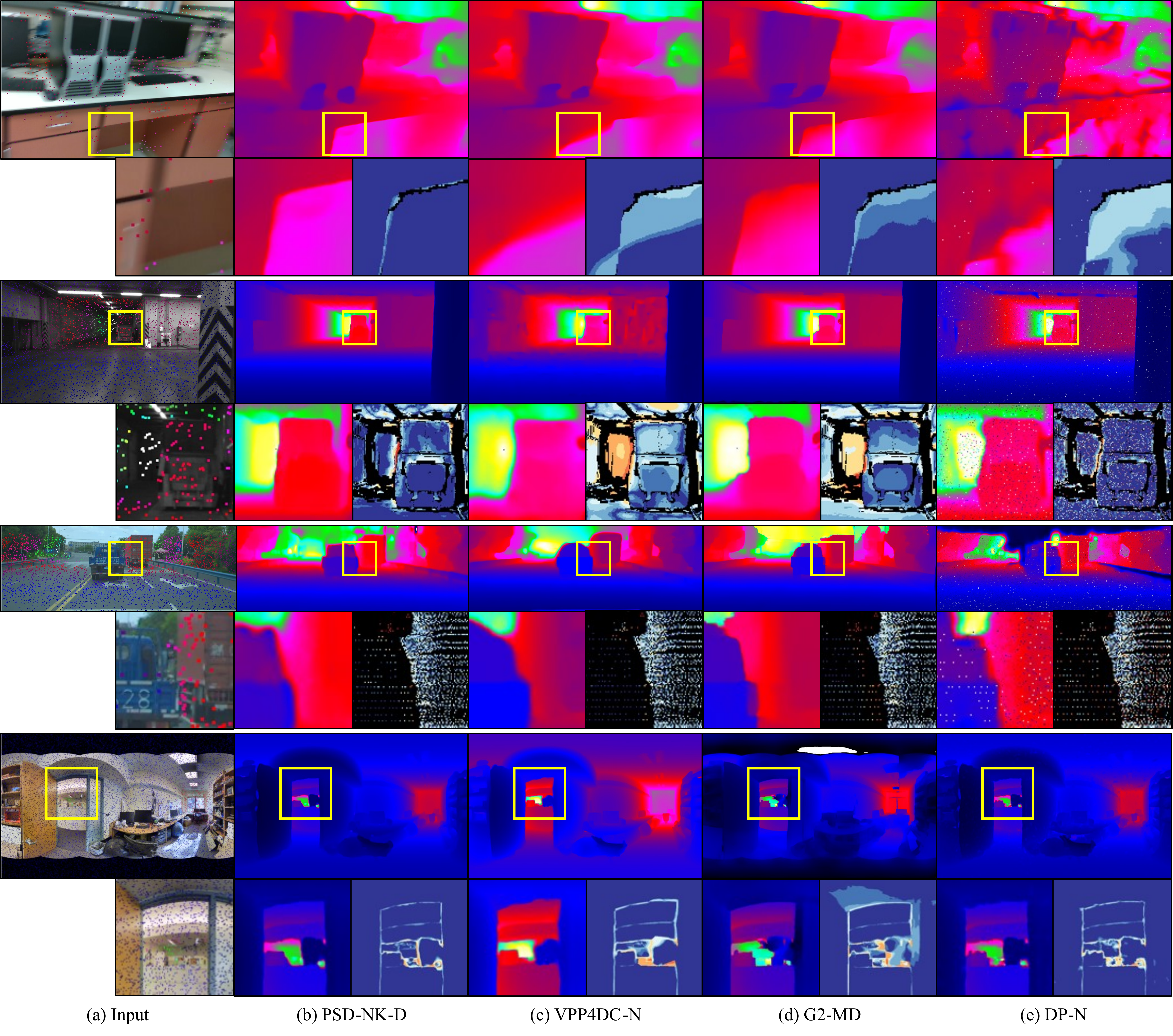}
\caption{\textbf{Qualitative results.} We compare our PSD-NK-D to VPP4DC-N \cite{bartolomei2023revisiting}, G2-MD \cite{10373158} and DP-N \cite{Park_2024_CVPR} on VOID \cite{wong2020unsupervised}, ETH3D \cite{8099755}, Drivingstereo \cite{yang2019drivingstereo} and Stanford2D3D \cite{ArmeniSZS17} datasets (\textit{from top to bottom}). For the boxed regions, we zoom in and show the error map (\textit{dark blue is better}).}
\label{fig_show2}
\end{figure*}

\begin{figure*}[t]
\centering
\includegraphics[width=1\linewidth]{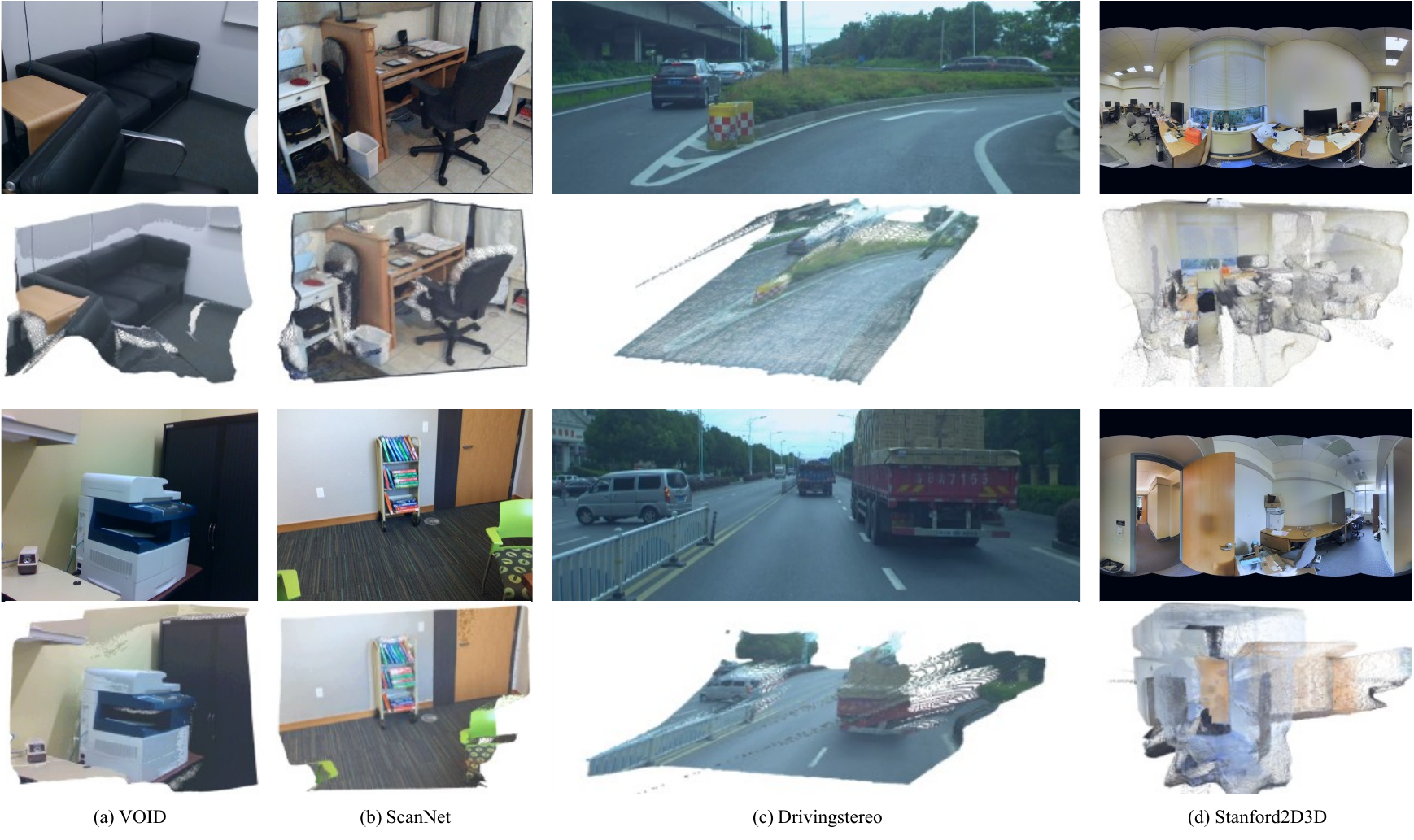}
\caption{\textbf{Scene Reconstruction.} We reconstruct the scene pointclouds (\textit{bottom}) using our PSD-NK-D on VOID \cite{wong2020unsupervised}, ScanNet \cite{8099744}, Drivingstereo \cite{yang2019drivingstereo}, and Stanford2D3D \cite{ArmeniSZS17}.}
\label{fig_show3}
\end{figure*}

\noindent
\textbf{Results on real-world datasets (without raw depth).}
TABLE \ref{tab_rate2} and TABLE \ref{tab_rate} report performance on real-world datasets without raw depth. To generate the sparse depth, we randomly sample 1\% of the pixels from the ground truth.
In indoor scenes, PSD gets competitive results across all datasets. Only VPP4DC-N \cite{bartolomei2023revisiting} is comparable to our PSD-N-D and PSD-NK-UDv2 on Middlebury \cite{ScharsteinS02}. Although CFormer-N \cite{zhang2023completionformer} shows the best performance on several metrics, its robustness on other datasets is suboptimal.
In outdoor scenes, our model also shows robust performance. On Argoverse \cite{8953693}, the MAE of PSD is significantly superior to other methods. However, PSD-K-D only exhibits marvelous robustness in outdoor scenes, which may be caused by the fact that the KITTI training data only contains the street scene that is overly similar to the test data, such as Drivingstereo \cite{yang2019drivingstereo}.

\noindent
\textbf{Results on synthesized datasets (without raw depth).}
As shown in TABLE \ref{tab_syn}, our framework exhibits significant generalization on two synthesized datasets. On Hypersim \cite{roberts2021hypersim}, the MAE of PSD-N/NK is far ahead of other methods. On VKITTI2 \cite{cabon2020vkitti2}, compared to the G2-MD \cite{10373158} trained on this dataset, PSD-K/NK-D achieves better results. 
Notably, both datasets are used to train Metricv2 \cite{HuYZCLCWYSS24}. However, PSD-NK-Mev2 outperforms other models only on Hypersim, suggesting that the depth foundation model is not the only factor determining overall performance. These results indicate that our framework successfully transfers the robustness of these models into depth completion.

\noindent
\textbf{Results on other camera models (without raw depth).} In TABLE \ref{tab_syn}, we compare the robustness of our proposed PSD with other methods on the KITTI 360 \cite{LiaoXG23} and Stanford2D3D \cite{ArmeniSZS17} datasets, which are captured using fisheye and panoramic cameras with distortion patterns differing from general pinhole cameras. While our PSD performs well on both datasets, it slightly underperforms compared to CFormer-N \cite{zhang2023completionformer} on Stanford2D3D \cite{ArmeniSZS17}. This might be due to the similarity between the NYUv2 training dataset and Stanford2D3D \cite{ArmeniSZS17}.

\noindent
\textbf{Visualization results.}
Fig. \ref{fig_show2} presents qualitative results on the VOID \cite{wong2020unsupervised}, ETH3D \cite{8099755}, Drivingstereo \cite{yang2019drivingstereo} and Stanford2D3D \cite{ArmeniSZS17} datasets. We show several results obtained from three robust methods: VPP4DC \cite{bartolomei2023revisiting}, G2-MD \cite{10373158}, and DP \cite{Park_2024_CVPR}. Compared to these methods, our PSD framework achieves better results with clear object boundaries. Fig. \ref{fig_show3} shows the scene reconstruction using our PSD. These demonstrate that our framework successfully integrates depth foundation models into depth completion, achieving remarkable robustness.

\noindent
\textbf{Comparison with other methods.}
Our framework achieves superior robustness compared to other methods in the OOD scenarios. Among them, CFormer \cite{zhang2023completionformer} is one of the state-of-the-art methods for specific scenes, yet its robustness is limited.
IP\_Basic \cite{ku2018defense} is a depth-only completion method that cannot restore the appearance of an object since the RGB image is ignored.
VPP4DC \cite{bartolomei2023revisiting} leverages a stereo matching network to achieve robust results.
G2-MD \cite{10373158} is trained on large-scale synthetic data, double of ours, and develops a strong data augmentation pipeline.
PromptDA \cite{abs-2412-14015} uses the Depth Anything v2 \cite{depth_anything_v2} for depth super-resolution. Similar to our framework, DP\cite{Park_2024_CVPR} also utilizes depth foundation models and propagation approaches. DP fuses the sparse depth measurements at the feature level via a complex network module, and performs propagation solely in 2D image space for post-processing, like CFormer \cite{zhang2023completionformer}.
In contrast, our PSD integrates sparse depth measurements at the depth level, leveraging structural and semantic contextual cues from the depth foundation model to propagate the sparse depth in both 3D and 2D spaces. Additionally, PSD includes an effective correction module to refine and correct distortions.

\noindent
\textbf{Performance of our components.}
Our ALIGN component and DS propagation component both show effectiveness on most datasets. Compared to CFormer \cite{zhang2023completionformer}, which is constrained to a single, homogeneous scene (outdoor or indoor), our components exhibit a remarkable robustness in both outdoor and indoor scenes, particularly the real-world datasets with raw depth. The success of ALIGN-D/Dv2/M indicates the effective alignment as a crucial factor in extending the depth foundation model to the depth completion task. More importantly, our DS propagation component is an effective approach for transferring the capability of these foundation models in representing appearance characteristics to address diverse depth characteristics. These results show that our DS propagation surpasses VPP4DC \cite{bartolomei2023revisiting}, G2-MD \cite{10373158}, and DP \cite{Park_2024_CVPR} on most datasets.

\begin{figure}[t]
\centering
\includegraphics[width=1\linewidth]{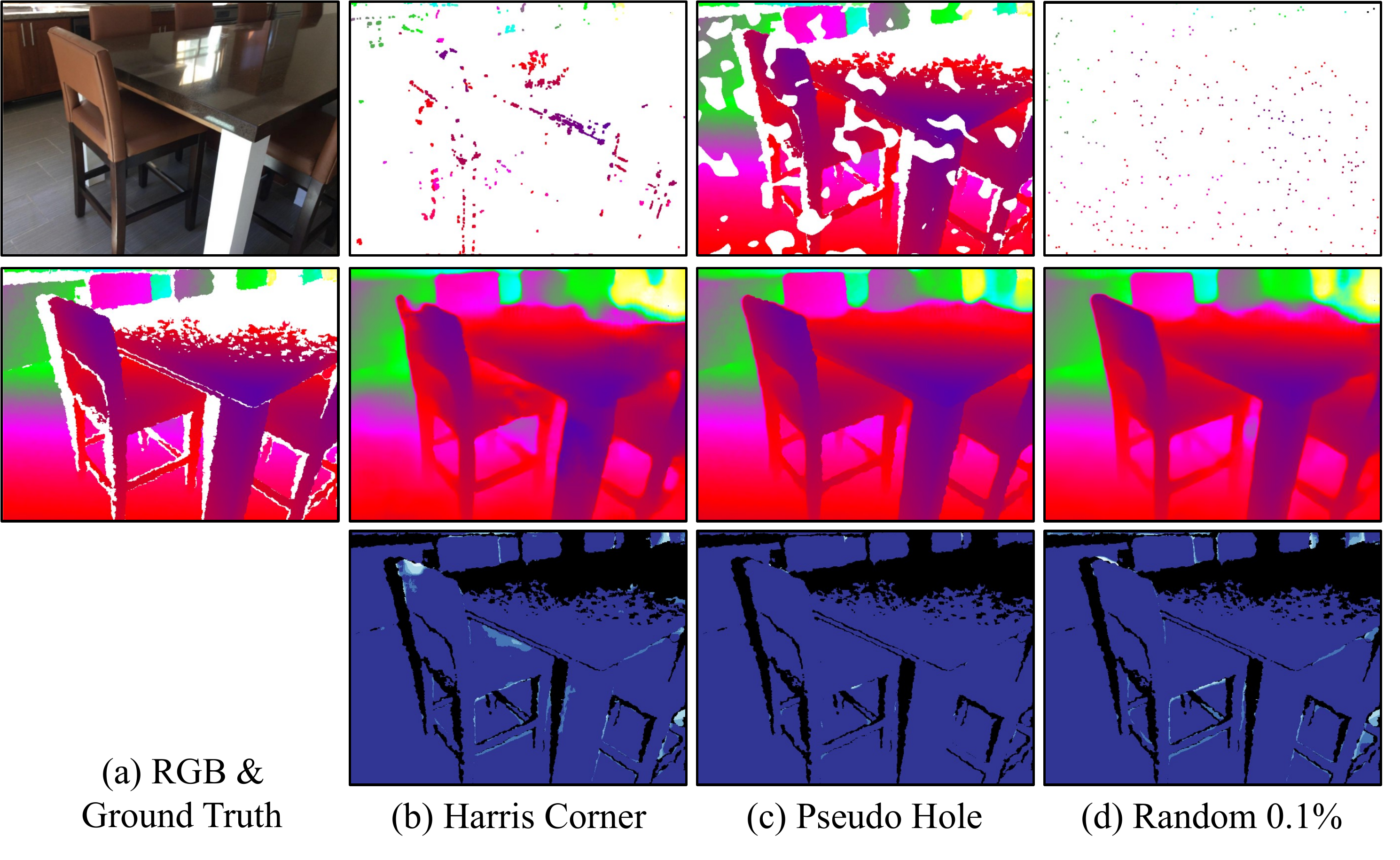}
\caption{{\bf Qualitative results at three sparse patterns.} In the first column, we show the RGB image and ground truth. For the other columns, we present the sparse depth, final prediction and error map at different patterns, respectively.}
\label{fig_pattern}
\end{figure}

\subsection{Robustness for Sparse Patterns}
To evaluate the robustness of our framework across different sparse patterns, we adopt 3 distinct patterns to generate the sparse depth, including Harris corner, pseudo hole and random mask (0.1\% sampling rate). As shown in Fig. \ref{fig_pattern}, these patterns pose significant challenges to most existing methods due to the substantial missing regions and heterogeneous distributions. Despite these challenges, our framework demonstrates remarkable robustness. This is primarily attributed to our dual-space propagation module, which can effectively diffuse sparse depth to surrounding regions regardless of the patterns.

\begin{figure*}[t]
\centering
\includegraphics[width=1\linewidth]{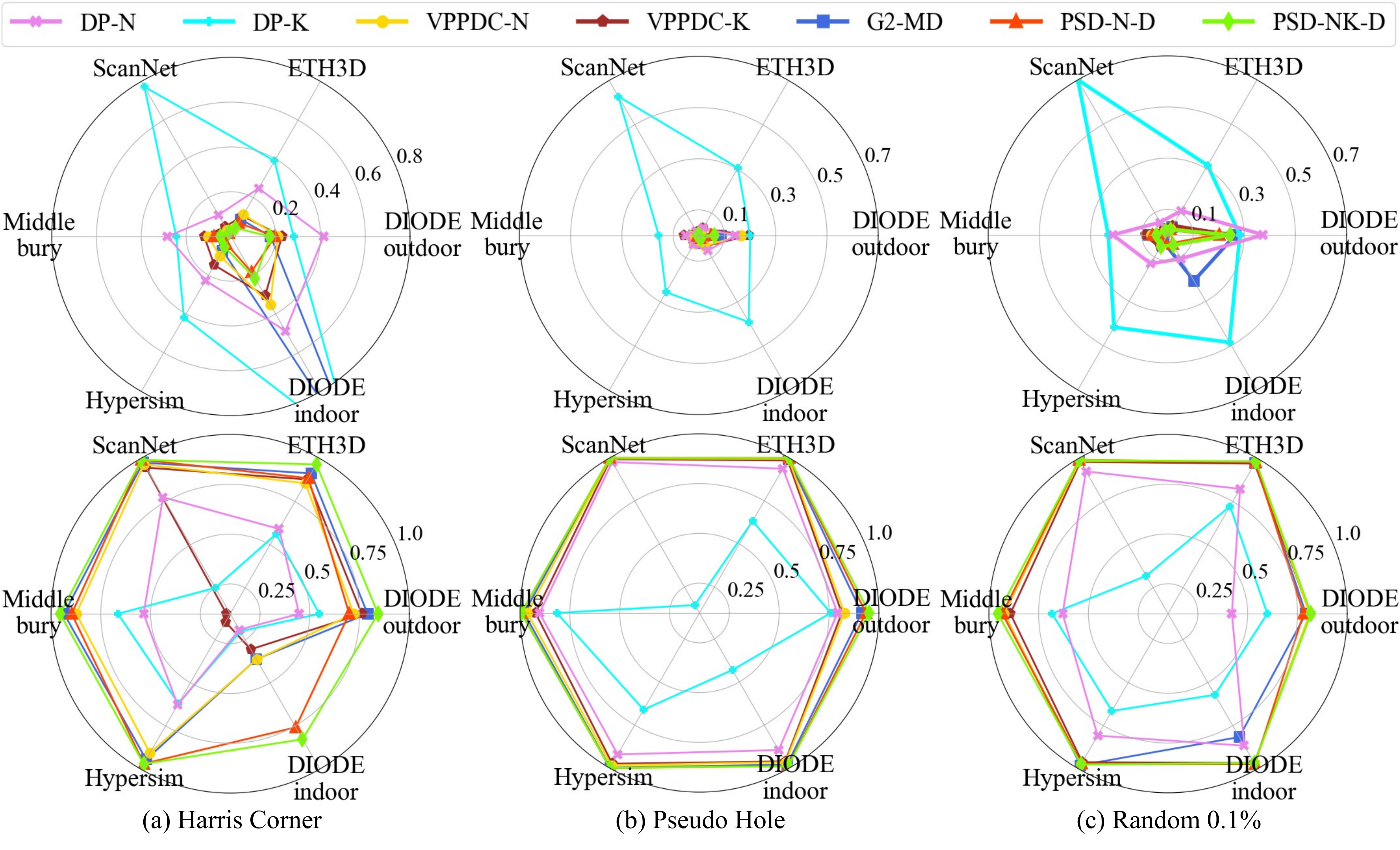}
\caption{\textbf{REL (\textit{top}) and $\mathbf{\delta_{1.25}}$ (\textit{bottom}) of different methods.} Across three sparse patterns on 6 datasets, the REL region of our PSD is smallest, and the $\delta_{1.25}$ region of our PSD is largest, indicating that our robustness is superior. DP-K \cite{Park_2024_CVPR} and G2-MD \cite{10373158} obtain invalid REL on the DIODE indoor \cite{vasiljevic2019diode} in the Harris corner pattern.}
\label{fig_radar}
\end{figure*}

Fig. \ref{fig_radar} shows quantitative results of our PSD and other generalizable methods, clearly demonstrating that the robustness of our model across three sparse patterns is superior on 6 different datasets. Despite the competitiveness demonstrated by other methods within the random mask (0.1\%), notable degradation is observed in the performance of them when evaluating on the other two patterns.
Our framework outperforms most methods across both the REL and $\delta_{1.25}$ metrics. While G2-MD \cite{10373158} exhibits competitiveness in terms of the $\delta_{1.25}$ metric, it fails to consistently obtain a valid REL value for the Harris corner evaluated on the DIODE indoor \cite{vasiljevic2019diode}.

\subsection{Ablation Study}
\noindent
\textbf{The effectiveness of depth foundation model.} The baseline is CSPN \cite{cheng2018depth}, a typical depth completion method that uses ResNet50 for feature extraction and employs 2D propagation to diffuse sparse depth measurements. In TABLE \ref{tab_cspn}, the first row shows its original results on NYUv2 \cite{SilbermanECCV12}/KITTI \cite{geiger2012we}, which is used for training. Next, we test its NYUv2 pre-trained model on KITTI that shows a significant performance drop. Then, we integrate Depth Anything \cite{depthanything} into CSPN \cite{cheng2018depth}. In this integrated approach, Depth Anything extracts robust features from RGB images, while CSPN serves as a parameter-free 2D propagation module to diffuse sparse depth. Furthermore, our DS propagation further improves performance. This experiment demonstrates the depth foundation model's effectiveness in enhancing the generalization of depth completion.

\begin{table}[t]
\centering
\caption{{\bf Effectiveness of the depth foundation model.} * denotes official results. DA denotes Depth Anything. DS denotes our propagation approach. Better: RMSE $\downarrow$, $\delta_{1.25}$ $\uparrow$.}
\begin{tabular}{l|c|cc|cc}
\hline
\multirow{2}{*}{Method} & \multirow{2}{*}{Training} & \multicolumn{2}{c|}{NYUv2 \cite{SilbermanECCV12}} & \multicolumn{2}{c}{KITTI \cite{geiger2012we}}\\ \cline{3-6}
& & RMSE    &  $\delta_{1.25}$ & RMSE  &  $\delta_{1.25}$ \\
\hline
CSPN \cite{cheng2018depth}* &NYUv2 or KITTI& \textbf{117.0} & \textbf{99.20} & 2977 & 95.70 \\ 
CSPN \cite{cheng2018depth} &NYUv2& 322.6 & 90.31 & 7796 &  8.749 \\ 
DA with CSPN 
&-&263.7&96.43&{1676}&\textbf{97.83}\\
DA with DS 
&-&161.3&98.41&\textbf{1662}&{97.77}\\
\hline
\end{tabular}
\label{tab_cspn}
\end{table}

\noindent
\textbf{Scale calibration.} 
The original scale-agnostic depth obtained from the depth foundation model belongs to the inverse depth space, a counterintuitive structure, which leads to instability and collapse if directly used as a result, as shown in TABLE \ref{tab_diff}. The least-squares based alignment (ALIGN ours) effectively maps the metric scale of the scale-agnostic depth, compared to alignments based on its reciprocal and medium value. To evaluate the effect of calibration on dual-space propagation, we directly used the scale-agnostic depth to determine neighbors and the initial value for comparison. The results demonstrate that scale calibration is the key to successfully embedding most depth foundation models in the propagation approach.

\begin{table}[t]
\centering
\scriptsize
\caption{{\bf Ablations on propagation modules.} The models are based on Depth Anything. 2D/3D denotes the space used for propagation. Better: MAE $\downarrow$, REL $\downarrow$, $\delta_{1.25}$ $\uparrow$.}
\begin{tabular}{l|ccc|ccc}
\hline
\multirow{2}{*}{Method} & \multicolumn{3}{c|}{NYUv2 \cite{SilbermanECCV12}} & \multicolumn{3}{c}{KITTI \cite{geiger2012we}}\\ \cline{2-7}
& MEA    & REL  & $\delta_{1.25}$ & MEA  &  REL  & $\delta_{1.25}$ \\ 
\hline
- 
&80994& - &1.564&101834& - &3.744  \\
Reciprocal
&1918924	& - &0.035&50718& - &0.001\\
ALIGN ours 
&\textbf{175.9}&\textbf{5.69}&\textbf{96.16}&\textbf{1970}&\textbf{9.56}&\textbf{91.05}\\
ALIGN medium 
&613.1&20.7&64.62&2861&14.3&86.30 \\
\hline
2D w/o alignment &49934&-&4.909&3465&42.4&63.92\\
3D w/o alignment &19232&-&14.06&74048&-&5.942\\
2D 
&137.9&4.46&96.43&510.5&3.09&97.83\\
3D 
&82.42&\textbf{2.88}&98.46&552.9&3.27&\textbf{98.94}\\
2D3D parallel
&96.55&3.19&98.37&\textbf{488.8}&2.96&98.42\\
2D-3D serial
&85.12&2.97&\textbf{98.47}&539.4&\textbf{2.25}&97.71\\
3D-2D serial
&\textbf{82.20}&2.93&98.41&511.5&3.12&97.78\\
\hline
\end{tabular}
\label{tab_diff}
\end{table}

\begin{figure*}[t]
	\centering
	\includegraphics[width=1\linewidth]{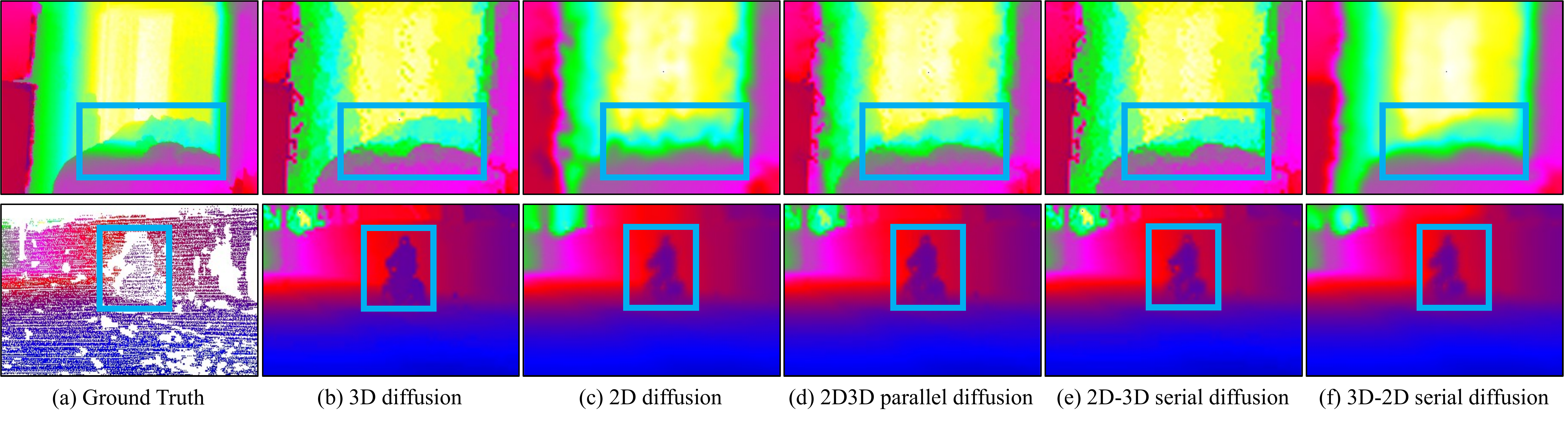}
	\caption{\textbf{Different propagation settings.} Top: NYUv2 \cite{SilbermanECCV12}. Bottom: KITTI \cite{geiger2012we}. We mark regions with boxes to highlight the results of different propagation settings.}
	\label{fig_2d3d}
\end{figure*}

\begin{table}[t]
	\centering
	\scriptsize
	\caption{{\bf Detailed choices in propagation module.} The models are based on Depth Anything, and we report the MAE for this ablation. M/P denotes Metric/Pre-filled depth.}
	\begin{tabular}{ccc|cccc|c|cc}
		\hline
		\multicolumn{3}{c|}{$k$} & \multicolumn{4}{c|}{$\eta$} & \multirow{2}{*}{Init} &\multirow{2}{*}{NYUv2 \cite{SilbermanECCV12}} & \multirow{2}{*}{KITTI \cite{geiger2012we}} \\ \cline{1-7}
		1   & 2   & 3  & 0.1 & 0.5 & 0.9 & 0.99 & &\\
		\hline
		\checkmark&          &    & & &\checkmark&     &M&101.2&1736 \\
		\checkmark&          &    & \checkmark& &&     &M&165.7&1943 \\
		\checkmark& &&\checkmark&     &     &          &P&147.5&561.7 \\
		\checkmark&&    &&\checkmark&     &            &P&104.3&\textbf{544.1} \\
		\checkmark&  &            & & &\checkmark&     &P&\textbf{82.42}&552.9 \\
		\checkmark&  &  &        &     &  &\checkmark  &P&83.91&559.6 \\
		&\checkmark&       &&     &\checkmark&         &P&85.48&552.9 \\
		&     &\checkmark    &&     &\checkmark     &  &P&89.80&553.0 \\
		\hline
	\end{tabular}
	\label{tab_3d}
\end{table}

\noindent
\textbf{Dual space \textit{vs.} Single space.} We compare our dual-space propagation with its single-space counterpart. The results in TABLE \ref{tab_diff} show that our proposed propagation approach based on 3D space is effective, even outperforming the existing propagation based on 2D space \cite{cheng2018depth} in certain scenes. As shown in Fig. \ref{fig_2d3d}, single-space propagation either emphasizes the 3D global spatial structure, resulting in clearer boundaries, or focuses on the local consistency of the 2D image space, which may lead to oversmoothing. Our dual-space propagation, regardless of the settings, propagates the sparse depth measurement in both 3D and 2D spaces, yielding better results.

\noindent
\textbf{Serial propagation \textit{vs.} Parallel propagation.} We study the setting of dual-space propagation, including 3D and 2D spaces. For the serial scheme, the process can be initiated with 3D space, followed by 2D space, or vice versa. For the parallel scheme, the sparse depth measurements are propagated independently in 3D and 2D spaces, and then the average of the results is taken as the final output. In Fig. \ref{fig_2d3d}, it can be observed that 2D-3D serial propagation generates clearer boundaries, whereas the 3D-2D version appears oversmoothed. On the other hand, the parallel version strives to find a balance between the two. As shown in TABLE \ref{tab_diff}, due to the variations in the appearance and depth distribution among diverse scenes, the same propagation approach is not suitable for all of them. 

\noindent
\textbf{The setting of 3D propagation.} Three key factors significantly influence the performance of propagation in 3D space: the number of neighbors, the initial value, and the updating coefficient. As shown in TABLE \ref{tab_3d}, the MAE deteriorates as the number of neighbors increases, since the spatial structure derived from the depth foundation model, which is merely a cue containing some distortions, does not perfectly match the real structure. However, given the diversity of OOD scenarios, we make 3D propagation user-configurable to promote the performance for specific scenes. 
In addition, during the propagation process, we only leverage metric depth to generate the spatial structure for searching for nearest neighbors. Since the metric depth is oversmoothed, the initial value is replaced with a pre-filled depth using Gaussian filters \cite{ku2018defense}. Finally, the updating coefficient $\eta$ is sensitive to the data, and we set it to 0.9 to evaluate the performance in all OOD scenarios.

\begin{figure}[t]
	\centering
	\includegraphics[width=1\linewidth]{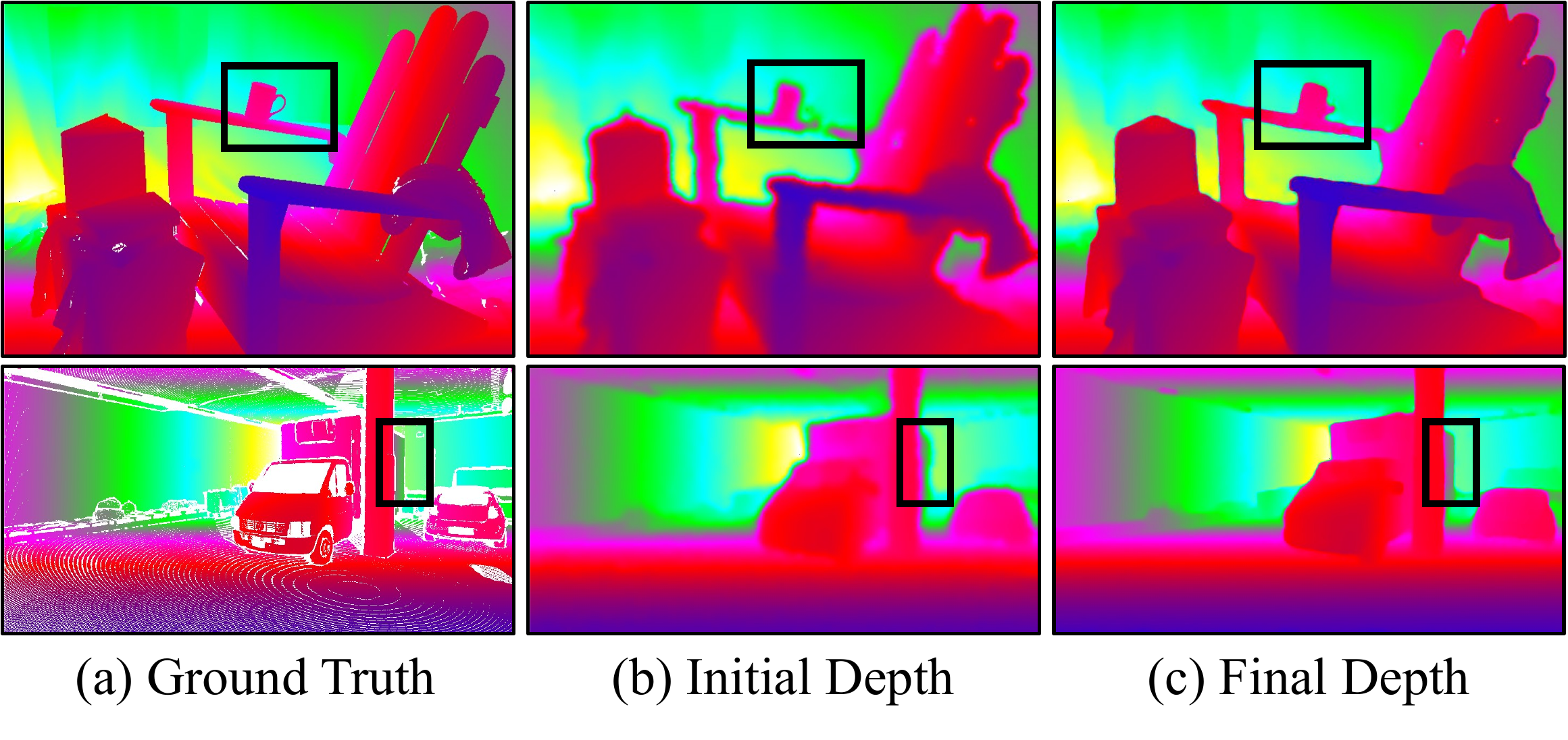}
	\caption{{\bf Effects of correction module.} Our correction module eliminates the structural distortion in the initial depth.}
	\label{fig_showcor}
\end{figure}

\noindent
\textbf{The effectiveness of correction.} Our progressive correction module comprises three stages: initial estimation, residual slice assignment and global offset adjustment. We gradually embed them based on our dual-space propagation. As shown in TABLE \ref{tab_cor}, these three stages prove effective for both training and unseen data.
The initial estimation effectively mitigates the residual range. In the subsequent stage, the performance is intimately tied to the number of residual slices. To balance the efficiency, we allocate 5 slices to divide the range for refinement. Finally, the global offset is used to continuous adjustment of slices, which is particularly effective for large residual range. Fig. \ref{fig_showcor} shows that our module effectively corrects the structural distortions present in the initial depth.

\begin{table}[t]
\centering
\scriptsize
\caption{{\bf Ablations on correction modules.} The models are based on Depth Anything and trained on NYUv2. Better: MAE $\downarrow$, REL $\downarrow$.}
\begin{tabular}{ccccc|cc|cc}
\hline
\multirow{2}{*}{Initial} & \multicolumn{3}{c}{Residual slice} & \multirow{2}{*}{Global offset} & \multicolumn{2}{c|}{NYUv2 \cite{SilbermanECCV12}} & \multicolumn{2}{c}{KITTI \cite{geiger2012we}} \\
& 3      & 5      & 7     &                    & MAE   & REL  & MAE  & REL  \\
\hline
-&-&-&-&-                         &79.89&2.89&512.3&3.12 \\
\checkmark&&&&                    &46.19&1.62&475.2&2.62 \\
\checkmark&\checkmark&&&          &42.27&1.48&464.9&2.80 \\
\checkmark&&\checkmark&&          &34.98&1.19&418.1&2.56 \\
\checkmark&&&\checkmark&          &36.14&1.24&430.4&2.58 \\
\checkmark&&\checkmark&&\checkmark&\textbf{34.84}&\textbf{1.18}&\textbf{393.4}&\textbf{2.47} \\
\hline 
\end{tabular}
\label{tab_cor}
\end{table}

\noindent
\textbf{Different uncertainty estimation.} In our PSD framework, we explicitly model uncertainty as the error between the predicted depth map and the ground-truth depth map (U1) \cite{10234651}. In TABLE \ref{tab_unc}, we compare this approach with another uncertainty estimation method (U2) \cite{9157454}, which is based on the concept of heteroscedastic aleatoric uncertainty in deep neural networks. Although both approaches effectively enhance depth refinement, U1 shows slightly better generalization ability.

\begin{table}[t]
\centering
\tiny
\caption{{\bf Comparison of different uncertainty estimation approaches.} * denotes training datasets. Better: MAE $\downarrow$, RMSE $\downarrow$, REL $\downarrow$, $\delta_{1.25}$ $\uparrow$.}
\begin{tabular}{l|cccc|cccc}
\hline
\multirow{2}{*}{Dataset} & \multicolumn{4}{c|}{U1} & \multicolumn{4}{c}{U2} \\
& MAE  & RMSE  & REL  & $\delta_{1.25}$  & MAE  & RMSE  & REL  & $\delta_{1.25}$ \\ \hline
NYUv2 \cite{SilbermanECCV12}*   &37.19&102.6&1.28&99.41&\textbf{35.87}&\textbf{100.4}&\textbf{1.22}&\textbf{99.42} \\
KITTI \cite{geiger2012we}* &\textbf{253.1}&\textbf{1218}&\textbf{1.43}&\textbf{99.44} &268.4&1244&1.54&99.34 \\ \hline
VOID \cite{wong2020unsupervised}&29.71&76.20&1.98&\textbf{99.02}&\textbf{28.25}&\textbf{75.14}&\textbf{1.85}&\textbf{99.02}  \\
ScanNet \cite{8099744}&\textbf{13.12}&\textbf{49.45}&\textbf{0.72}&\textbf{99.62}&\textbf{13.12}&50.54&0.76&{99.55} \\
ETH3D \cite{8099755}& 29.27&172.1&\textbf{0.64}&\textbf{99.63}&\textbf{28.66}&\textbf{159.1}&0.67&{99.58}\\
Hypersim \cite{roberts2021hypersim}&77.84&389.8&\textbf{2.47}&\textbf{98.69}&\textbf{76.32}&\textbf{372.1}&4.37&{98.52}  \\
DrivingStereo \cite{yang2019drivingstereo}&\textbf{556.3}&\textbf{1655}&\textbf{1.78}&\textbf{99.45}&560.8&1661&1.81&99.43  \\
Cityscapes \cite{7780719} & 2471&\textbf{10803}&\textbf{11.39}&{97.15}&\textbf{2450}&10920&12.47&\textbf{97.27}\\ \hline
\end{tabular}
\label{tab_unc}
\end{table}

\subsection{Discussion}
\noindent
\textbf{Factors about robustness.} We discover that there are two factors affecting the robustness of depth completion: appearance characteristics from RGB images and depth characteristics from sparse depth maps. 
Recent foundation models \cite{9178977,depthanything,depth_anything_v2} for monocular depth estimation have achieved exceptional robustness by collecting large-scale training data with diverse appearance characteristics. Our PSD framework successfully leverages the depth foundation model to tackle the challenge posed by various appearances, and proposes a propagation approach to transfer this capability into the depth domain in order to address diverse depth characteristics. Thus, the robustness of our framework primarily stems from depth foundation models, and our designs successfully transfer their robustness to depth completion.

\noindent
\textbf{Training data.} 
We use the NYUv2 \cite{SilbermanECCV12} and KITTI \cite{geiger2012we} datasets to train the proposed framework. 
While our PSD-N/NK model exhibits robustness to most OOD scenarios, the PSD-K model only performs well on outdoor scenes, since KITTI and datasets like Drivingstereo \cite{yang2019drivingstereo} share excessively similar appearance and depth characteristics. In addition, VPP4DC-N/K \cite{bartolomei2023revisiting} and DP-N/K \cite{Park_2024_CVPR} show this same phenomenon.
Although KITTI is larger in scale than NYUv2, it is collected primarily from urban street, resulting in a more uniform data pattern. In contrast, NYUv2 provides a more diverse and richer set of scenarios, encompassing a wider range of visual contexts. 
This confirms the success factors of previous foundation models, enhancing the diversity and scale of data is crucial for model robustness.


\noindent
\textbf{Limitation.} 
While our framework has yielded satisfactory results, two problems remain unresolved.
First, PSD is limited to the depth completion task, due to the propagation approach, which prevents it from being a unified framework for depth inference. When sparse depth information is excessively available, the framework exhibits low efficiency, and it cannot work properly without sparse depth.
Second, our PSD is trained on two typical datasets, NYUv2 \cite{SilbermanECCV12} and KITTI \cite{geiger2012we}. The scale and diversity of both datasets are considerably less than those of other datasets commonly used in training foundation models. If the scale and diversity of the training data are augmented, there would be a potential for further improving the performance of PSD in OOD scenarios. Given that most public RGB-D datasets in the real-world do not provide sufficient raw depth maps for training, a semi-supervised approach may be employed to address this problem \cite{depthanything}.

\noindent
\textbf{Influence of foundation models.}
Foundation models such as GPT \cite{radford2018improving}, CLIP \cite{RadfordKHRGASAM21}, and DINO \cite{OquabDMVSKFHMEA24} have revolutionized the computer vision community due to their exceptional feature representation capabilities. Furthermore, leveraging these models to enhance robustness/generalization has become a dominant trend. For example, OVSeg \cite{LiangWDLZ0ZVM23}, PointCLIP \cite{9878980}, DenseCLIP \cite{9878572}, and Depth Anything \cite{depthanything} utilize these models to improve robust segmentation, detection, and depth estimation. In this work, we employ depth foundation models like Depth Anything to boost the generalization of depth completion in OOD scenarios, expanding their practical applications. Robots equipped with depth sensors such as structured light and time-of-flight can utilize our framework to complete sparse depth maps, improving performance across diverse scenarios.

\section{Conclusions}
In this paper, we propose a novel depth completion framework, named PSD, which possesses remarkable robustness in handling unseen data from the OOD scenarios. 
We leverage the existing depth foundation model to guide the propagation of sparse depth into missing regions to reconstruct a dense depth map.
Furthermore, we design a dual-space propagation approach that propagates sparse depth in the 3D Euclidean space and 2D image space to improve the 3D structure and local consistency. Without requiring any additional training or fine-tuning, our propagation approach achieves comparable robustness to most existing methods.
In addition, we introduce a learnable correction module that aligns the predicted depth with the ground truth to correct the disrupted structure caused by the distortions of the foundation model. Extensive experiments on 16 public datasets demonstrate that our PSD surpasses existing state-of-the-art depth completion methods in the OOD scenarios.

In future work, we will focus on two problems to further improve OOD performance of depth completion. First, our PSD is a framework specifically designed for depth completion. For general depth inference, a generative framework may be employed to address the task with multiple conditions \cite{Ke_2024_CVPR}. Second, our training dataset faces the limitations of scale and diversity. A semi-supervised approach may be adopted to address this problem \cite{depthanything}.

{\small
\bibliographystyle{ieee_fullname}
\bibliography{ref}
}

\end{document}